# Determining *Leishmania* Infection Levels by Automatic Analysis of Microscopy Images

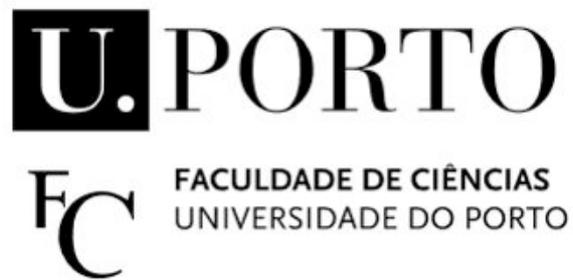

Pedro Gonçalo Ferreira Alves Nogueira

# Determining *Leishmania* Infection Levels by Automatic Analysis of Microscopy Images

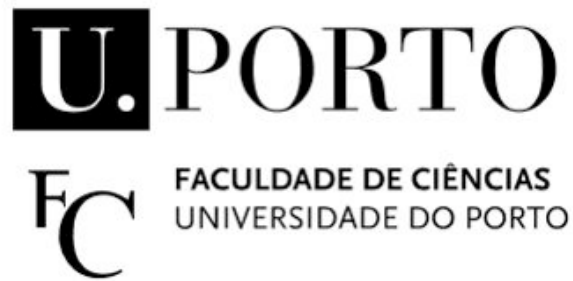

Pedro Gonçalo Ferreira Alves Nogueira

*"Beneath this mask there is an idea, and ideas are bulletproof."*

*- V*

# Acknowledgments

Firstly, I would like to thank my advisor, Prof. Miguel Coimbra, for all the opportunities and challenges he provided, for the continued support, for his patience, enthusiasm, availability and last but not least, for inciting me to push ever further and grow. In sum, I'd like to offer my most honest gratitude for helping me come this far.

To all my friends (which I am sorry to not mention in detail) for listening to my enthusiastic and incessant rants about my hypothesis, concept divagation and findings related to this work. I'd also like to thank all of them for the useful discussions and suggestions.

Lastly, I'd like to thank everyone at the IBMC/INEB laboratory for their full cooperation and enthusiasm and also my friend Luís, for his help.

This work was funded by the CellNote IT/LA/1086/2011 project coordinated by Instituto de Telecomunicações – Universidade do Porto.

# Abstract

Analysis of microscopy images is one important tool in many fields of biomedical research, as it allows the quantification of a multitude of parameters at the cellular level. However, manual counting of these images is both tiring and unreliable and ultimately very time-consuming for biomedical researchers. Not only does this slow down the overall research process, it also introduces counting errors due to a lack of objectivity and consistency inherent to the researchers' own human nature.

This thesis addresses this issue by automatically determining infection indexes of macrophages infected by the *Leishmania* parasite in microscopy images using computer vision and pattern recognition methodologies. Initially images are submitted to a pre-processing stage that consists in a normalization of illumination conditions. Three algorithms are then applied in parallel to each image. Algorithm A intends to detect macrophage nuclei and consists of segmentation via adaptive multi-threshold, and classification of resulting regions using a set of collected features. Algorithm B intends to detect parasites and is similar to Algorithm A but the adaptive multi-threshold is parameterized with a different constraints vector. Algorithm C intends to detect the macrophages' and parasites' cytoplasm and consists of a cut-off version of the previous two algorithms, where the classification step is skipped. Regions with multiple nuclei or parasites are processed by a voting system that employs both a Support Vector Machine and a set of region features for determining the number of objects present in each region. The previous vote is then taken into account as the number of mixtures to be used in a Gaussian Mixture Model to *decluster* the said region. Finally each parasite is assigned to, at most, a single macrophage using minimum Euclidean distance to a cell's nucleus, thus quantifying Leishmania infection levels.

The software framework was implemented in Java, using Weka's external libraries to train the Support Vector Machine and perform the Gaussian Mixture analysis. We were able to count macrophages and parasites with reasonably high accuracies (above 90%), and *decluster* regions with multiple nuclei or parasites with 75-85% accuracy. Ultimately, our results show our approach is able to replace a human in this task, alas with some room for improvement.

# Resumo

A análise de imagens de microscopia constitui uma ferramenta fundamental no campo da investigação biomédica, uma vez que permite a quantificação de uma multitude de parâmetros a nível celular. No entanto, a contagem manual destas imagens é cansativo e pouco confiável, sendo, no final de contas, muito demorada para os investigadores de ciências biomédicas. Este processo de contagem manual não só atrasa todo o processo de investigação, como introduz erros nas próprias contagens devido a uma falta de objectividade e consistência inerentes à condição humana dos investigadores.

Nesta tese propomos uma solução que visa determinar automaticamente os níveis de infecção em imagens de microscopia infectadas com o parasita *Leishmania*. Para este efeito são usadas técnicas de visão computacional e reconhecimento de padrões. As imagens são submetidas a um passo de pré-processamento que consiste na normalização das condições de iluminação. Três algoritmos diferentes são então aplicados em paralelo a cada imagem. O algoritmo A pretende detectar núcleos dos macrófagos e consiste numa segmentação por via de um *threshold* múltiplo adaptativo e na classificação das regiões resultantes, através de um conjunto de características. O algoritmo B pretende detectar parasitas e é semelhante ao algoritmo A, sendo o *threshold* múltiplo adaptativo parametrizado com um vector de restrições diferente. O algoritmo C tem como objectivo detectar o citoplasma presente na imagem, consistindo numa versão mais curta dos dois algoritmos anteriores, no qual não é feita qualquer classificação das regiões resultantes da segmentação. Regiões com múltiplos núcleos de macrófagos ou parasitas são processadas por um sistema de voto que determina o número de objectos presentes através de uma Support Vector Machine e algumas características da região. O voto anterior é então usado como o número de misturas a usar num Modelo de Misturas Gaussianas, que separa a região em macrófagos (ou parasitas) individuais. Finalmente, cada parasita é associado a, no máximo, um macrófago usando a distância Euclideana mínima ao seu centro, quantificando assim os níveis de infecção.

A *framework* foi implementada em Java, com recuso às bibliotecas externas da Weka para o treino da SVM e a análise das misturas gaussianas. Conseguimos contar células e parasitas com altas taxas de precisão (acima dos 90%) e separar conjuntos de regiões com 75 a 85% de precisão. Numa última análise os resultados mostram que a nossa abordagem consegue substituir um humano na tarefa em questão, havendo ainda algum lugar para melhorias.

# Table of Contents

# List of Tables

# List of Figures

# List of Abbreviations

**FMI – Fluorescence Microscopy Imaging**

**GMM – Gaussian Mixture Model**

**EM – Expectation Maximization**

**ML – Machine Learning**

**AI – Artificial Intelligence**

**CCA – Connected Component Analysis**

**CCL – Connected Component Labelling**

**FCC – Freeman Chain Code**

**MMB – Minimum Bounding Box**

**PR – Pattern Recognition**

**ANN – Artificial Neural Network**

**FFNN – Feed-Forward Neural Network**

**SVM – Support Vector Machine**

**WEKA – Waikato Environment for Knowledge Analysis**

**GUI – Graphics User Interface**

**ROI – Regions Of Interest**

**LLF – Low-Level Features**

**SNR – Signal to Noise Ratio**

**LL – Log-Likelihood**

# Chapter I

# Introduction

*Leishmania* is a genus of Trypanosomatid protozoa, and is the parasite responsible for the disease leishmaniasis [1]. It spreads through sandflies and its primary hosts are vertebrates, mainly rodents, canids and humans in developed countries. The parasite currently affects 12 million people throughout 88 countries [2]. Leishmaniose has two main different forms:

- Cutaneous leishmaniasis, which affects the skin and mucus membranes, forming sores which can take years to heal;
- Systemic (or visceral) leishmaniasis, which affects the entire body. This form can lead to death as the parasites are able to severely damage the immune system, reducing its ability to fight off the disease;

In cutaneous leishmaniasis symptoms include skin sores, ulcers and erosion of the mouth, tongue, gums, lips and nose, nosebleeds, breathing and swallowing difficulty. Systemic leishmaniasis not only deforms the host, as its cutaneous counterpart, but it also lays waste to its internal organs. Symptoms include abdominal discomfort, weeklong fever cycles, night sweats, scaly, grey, dark and ashen skin, thinning hair, fatigue and weight loss. Figure 1:1 shows a few examples of individuals affected by either forms of the disease, some in extremely advanced stages of the infection [3].

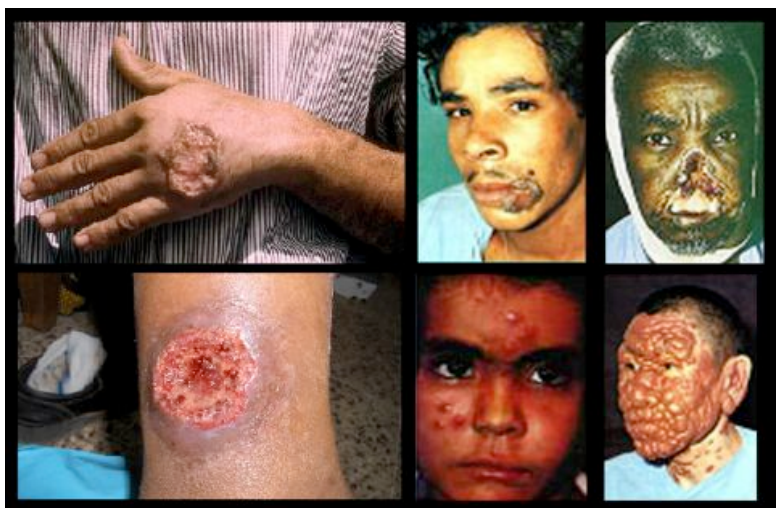

Figure 1:1 - Leishmaniose patients in various stages of the disease[1].

*Leishmania* is treatable by chemotherapeutics, which, nevertheless, suffer from poor administration regimen and host toxicity. Moreover, there are also cases of drug resistance to the currently available leshmanicidal drugs. However, in most cases plastic surgery is needed to correct the disfigurement caused by sores. Additionally, some patients need to have their spleen removed due to drug resistance. Although the disease is not generally deadly, it severely damages the immune system, leaving the body exposed to other deadly pathogens [3]. The inadequate means to treat leishmaniasis render the research for new medicines an urgent task.

Research in *Leishmania* and related parasites produces large amounts of microscopy images which, in turn, require large amounts of time to classify and annotate. In a single laboratory the number can easily ascend to the thousands of images with just a dozen of different experiments (as each experiment normally produces from 100 to 180 images). Not only does this detract the researchers from exploring new alternatives by robbing them of useful time, as it is also prone to inter-person variance. This happens because, although there are conventions on the annotation process, many images are extremely complex and this leads to different final counts. Furthermore, the time consuming (and therefore mentally straining) nature of the annotation process means there is intra-person variance, which expresses itself as a decaying function over time as the subject gets tired, frustrated, bored or pressured to finish each annotation. All of the aforementioned reasons lead to the need for automatic or semi-automatic methods for image classification and annotation. These involve not only segmentation, but also region classification and the subsequent annotation process itself, in a consistent pipeline.

---

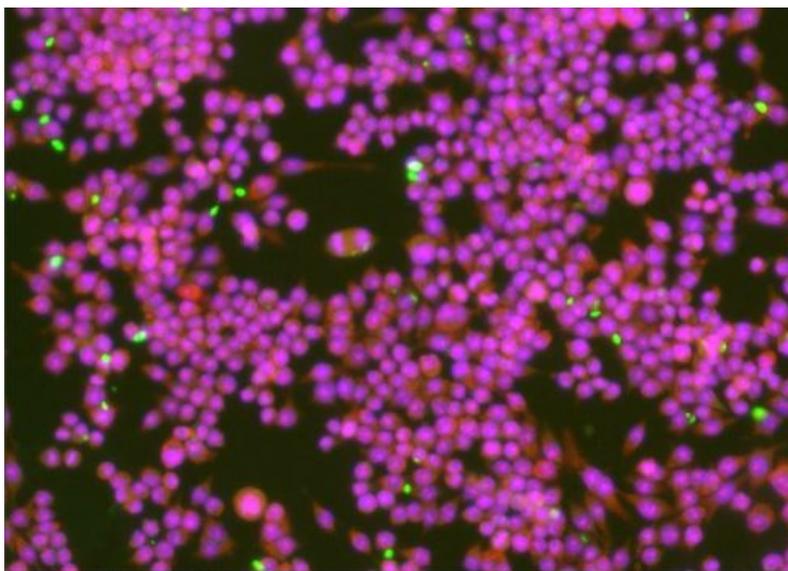

Figure 1:2 - A highly cluttered *Leishmania* microscopy image. Pink: cells; Green: parasites and ill-marked DNA.

To the best of our knowledge, no work has been developed in order to create a unifying pipeline for cell detection/counting and automatic annotation. All of the commercial software designed to aid in biomedical research focuses on cell detection, but does not provide any kind of automatic declustering or cell-parasite association, consisting mainly of simple generic computer vision algorithms that can be performed in a sequence and fine-tuned by the user. Although this is helpful it also lacks a sense of direction and purpose, posing as a set of tools rather than a (even if partial) solution. For further information on this subject we refer the reader to sub-section 2.6 of chapter 3.

Also, most of the new proposed methods focus on the segmentation step. Some of the most relevant approaches are the following: Comaniciu et al. [4] detects the cell clusters in the LUV colour space and delineates their borders by employing the gradient ascent mean shift method. In [5] the segmentation is proposed through a Canny edge detector, followed by a circle detection algorithm. An interesting approach in [6] proposes a hierarchical thresholding scheme using a priori information regarding the chromatic properties of background and cell classes. This last approach inspired us to look for similar patterns in *Leishmania* images, which later then led to the first part of our segmentation step. Sinha et al. [7] introduced a method for blood cell segmentation using the EM algorithm. This approach, while requiring no interaction from the user has a limited usability when faced with clustered cells. For a more extensive discussion of the related work please refer to section 3.1 - State of the Art.

In computer vision there is generally a trade-off between robustness and flexibility, which leads to either generic approaches that do not provide a great deal of detailed information, or to expert systems that while accurate in their designated task do not deal well when outside the scope of this task or when faced with noise (or outliers). Most techniques are sensitive to the right selection of parameters such as, threshold, mask-size and initial contour, not to mention acquisition conditions. Also, certain assumptions, such as, e.g. that a cell has a circular shape, are unviable for almost all cases of abnormal or clustered cells.

In this thesis we present a robust method for cell and parasite detection and counting, leading to the quantification of *Leishmania* infection levels. The cells are segmented through the usage of a priori information on the colour distribution found in these images as constraints for an adaptive multi-threshold algorithm. We then expand on the work in [7] by classifying the segmented regions through statistical parameters and allowing the usage of the EM algorithm as a means to support declustering of complex clustered regions through a linear voting system. Our method also has the advantage of being free from the constraints present in most of the related work and need of user interaction to fine-tune parameters which, most of the time, are comprehensible only to an expert.

This thesis aims at presenting four different contributions, which are the following:

I) Creation of a dataset for the testing and development of solutions for the automatic annotation of *Leishmania*-infected microscopy images;

II) A solution for this problem through the use of segmentation and classification techniques;

III) Refinement of the initial solution through the introduction of a voting system based on statistical classification and machine learning algorithms;

IV) Creation of a standard pipeline for cellular image processing (which can be extended and adapted to other scenarios due to its modularity);

Our solution also has the advantage of being easily integrated into our CellNote platform [8], which is currently deployed in three laboratories and used in a daily basis. Additionally, an early version of the final results was published in the proceedings of the Congress on Numerical Methods in Engineering. We expect to publish the final architecture and results in the near future.

## 1.1   Materials

For this study 794 fluorescence microscopy images were collected and used. These images were collected through a light microscope and annotated manually by a *Leishmania* research team at INEB/IBMC, led by Prof. Ana Tomás.

Fluorescent images are captured by staining the samples with different reagents (fluorescent molecules called fluorophores), which bind specific components of the cells under study (both *Leishmania* parasites and macrophages). When the sample is bombarded with light of a specific wavelength (or wavelengths) it's absorbed by the fluorophores, causing them to emit light of longer wavelengths (i.e. of a different colour than the absorbed light) [9]. This process is illustrated in Figure 1.1:1. For a more detailed description of the used materials, please refer to Chapter V – Results and Discussion, sub-section 1.

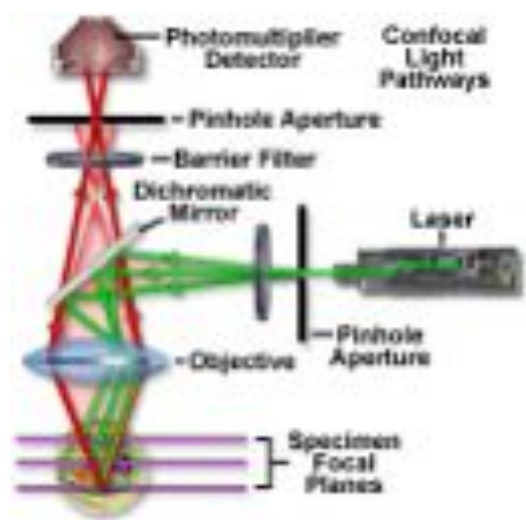

Figure 1.1:1 - Structure of a fluorescence microscope[2].

In the images used in this study three fluorophores with different colours were used, which allowed us to distinguish the nucleus (blue and red) and the cytoplasm (red) of the infected cells, as well the parasite as a whole (green). This provided us with three separate sets of data per image (one in each colour channel), motivating the treatment of cells, parasites and cytoplasm as individual images (Figure 2.2.1:2).

Specialized personnel individually annotated these images, with our CellNote software. This provided us with a solid ground truth for result evaluation.

---

## 1.2   Organization

The remainder of this thesis is organized as follows. In chapter II we describe the related work. This chapter is divided into two sub-sections; section 2.1 discusses the relevant state of the art, its advantages and disadvantages and compares those approaches to ours; section 2.2 provides the reader with the needed background for understanding the developed work, as well as with the reasons for the choice of each used method. Chapter III aims at providing the basics of microscopy image acquisition, reactants, different types of images and observed types of noise. In chapter IV, the system implementation is presented. This chapter is divided into seven sub-sections, each representing one of the algorithm's stages. Chapter V describes and discusses the contributions derived from the presented work, also focusing on their applicability to other similar problems. In chapter VI the benchmark results are presented and discussed, as well as their implications and repercussions on some initial hypothesis. Finally, chapter VII draws the final remarks on the presented work and obtained results, while commenting on the (currently on-going) future work.

# Chapter II

# Microscopy Imaging

In this chapter a brief introduction to microscopy imaging is given. We start by describing the principals and architecture of traditional optical techniques, followed by an also brief introduction to fluorescence and fluorescence microscopes. We conclude by discussing the most relevant challenges presented by the use of this type of images.

## 2.1 Optical Microscopy

Optical (or light) microscopes are the most basic type of magnifying devices available and first emerged, around the beginning of the seventeenth century.

These microscopes use light waves that are bent as they travel through a piece of convex glass. The idea is to bend diverging (spreading-out) light into a parallel path and then bend that parallel-path light into a small focus at the eyepiece. The simplest type of optical microscope is called a single lens microscope and, as the name implies, it uses only one lens for magnification. The simplest example of this kind of microscope would be a common magnifying class. Compound microscopes use systems of multiple lenses to focus the light into the eye or a camera. This kind of microscope is usually found in professional settings, as they are much more expensive due to the increased number of lenses and complex systems to keep them synchronized. An example of such system can be viewed in Figure 2.1:1. Optical microscopes are, however, limited to a magnification scale of nearly 1,000 times, even with multiple lens systems [10].

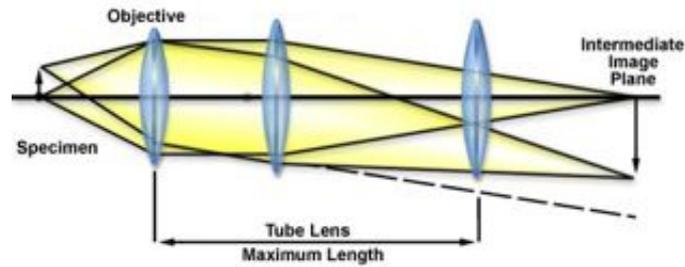

Figure 2.1:1 - The optical path in an optical microscope[3]. The objective lens at the end of the microscope tube is brought very close to the sample (because the lens have a very short focal length) and it is enlarged through a series of precisely aligned subsequent magnification lens.

A typical optical microscope has the same fundamental structural components [10]. These are listed bellow:

- Ocular lens (eyepiece): used to focus and observe the sample;

- Objective turret: used to rotate between multiple (final) lenses;

- Objective: which provides the first magnification of the sample;

- Stage focus wheel: moves the stage closer of farther away from the objective;

- Stage: to place the sample;

- Light source: to illuminate the sample (usually a light or a mirror);

These elements can also be observed in Figure 2.1:2, which represents a compound microscope that uses environmental light to irradiate the sample.

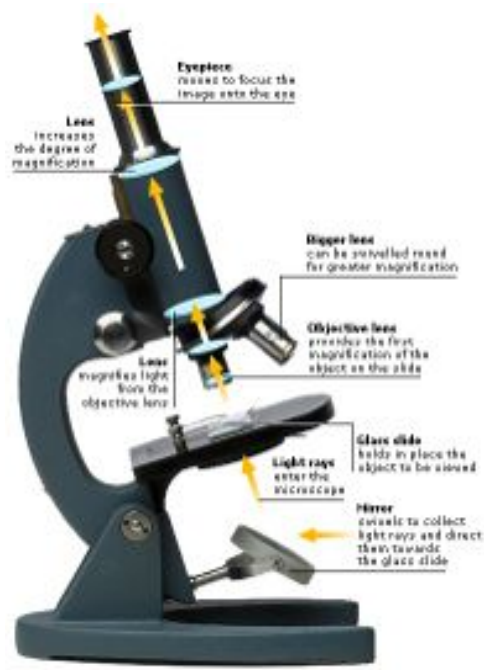

Figure 2.1:2 - Optical microscope architecture[4].

## 2.2 Introduction to Fluorescence Microscopy

Fluorescence is a member of the luminescence family of processes in which susceptible molecules emit light from excited states created by either a physical (e.g. absorption of light), mechanical (e.g. friction), or chemical mechanism.

The fluorescence process is governed by three important events, all of which occur on timescales that are separated by several orders of magnitude. Excitation of a susceptible molecule by an incoming photon happens in femtoseconds ($10^{-15}$ seconds), while vibrational relaxation of excited state electrons to the lowest energy level is much slower and can be measured in picoseconds ($10^{-12}$ seconds). The final process, emission of a longer wavelength photon and return of the molecule to the ground state, occurs in a time period of nanoseconds ($10^{-9}$ seconds). Because of the tremendously sensitive emission profiles, spatial resolution, and high specificity of fluorescence investigations, the technique has rapidly become an important tool in genetics and cell biology [11].

The category of molecules capable of undergoing electronic transitions that ultimately result in fluorescence is known as fluorescent probes or dyes, **fluorochromes**, or **fluorophores**. Fluorophores are divided into two broad classes, termed **intrinsic** and **extrinsic**. Intrinsic fluorophores, such as aromatic amino acids and neurotransmitters, and green fluorescent protein, are those that occur naturally. Extrinsic fluorophores are synthetic dyes or modified biochemicals that are added to a specimen to produce fluorescence with specific spectral properties [11]. In the material used for this work, extrinsic fluorophores were used [8].

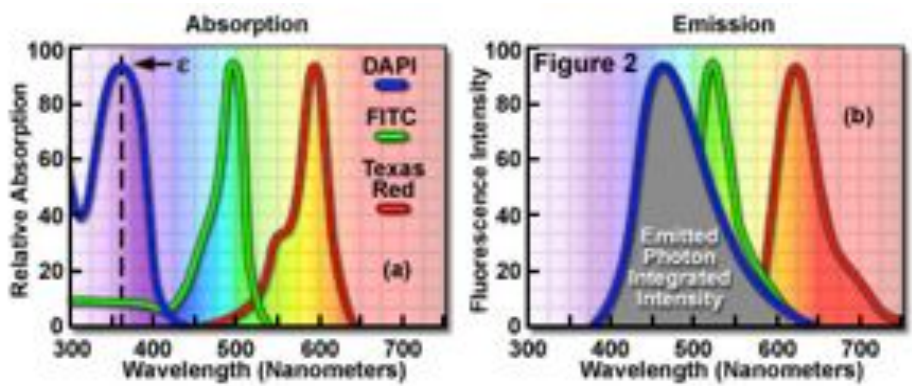

Figure 2.2:1 - Spectral profiles of two popular traditional fluorophores[5].

## 2.2.1 Fluorescence Microscopy

In contrast to the classical optical microscopy, the use of fluorescence microscopy allows the simultaneous labeling of different cell components, which can be easily distinguished based on the fluorescence properties of their specific dyes.

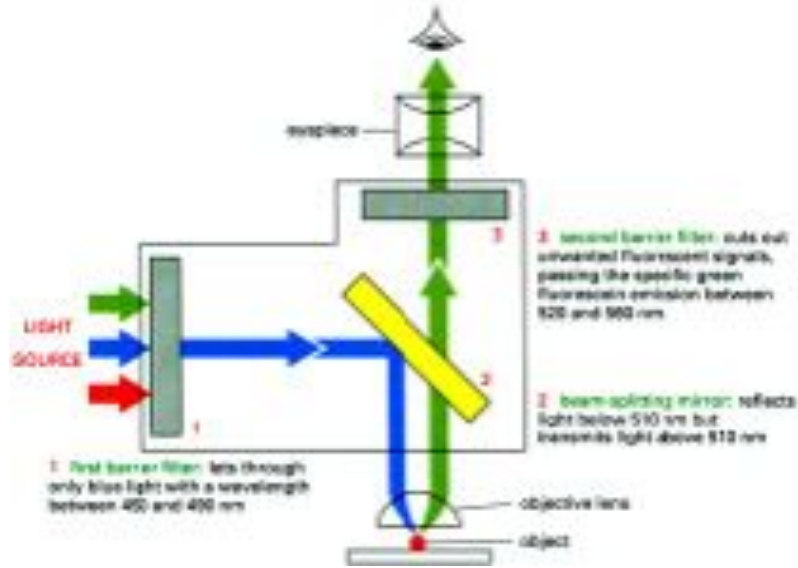

Figure 2.2.1:1 - Architecture of a generic fluorescence microscope [9].

The images collected for this study used three fluorophores, which emitted three distinct wavelengths. These allowed us to distinguish the nucleus (in blue), the cytoplasm (in red) of the infected cells and the parasites as a whole (in green). This provided three separate sets of data per image (one in each colour channel), motivating the treatment of cells, parasites and cytoplasm as individual images.

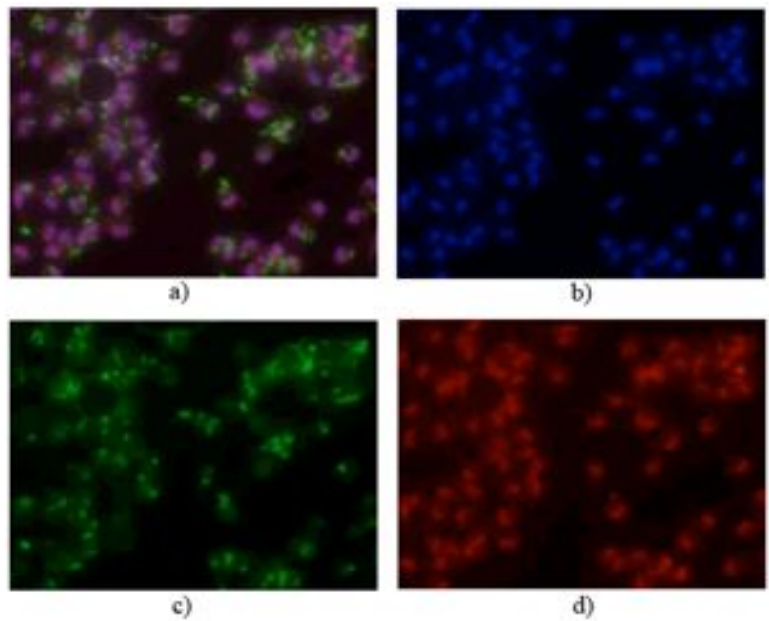

Figure 2.2.1:2 – Details of a fluorescence microscopy image. a) original image; b-c-d) macrophage nuclei, parasite nuclei and cytoplasm + macrophage and cell nuclei channels, respectively.

## 2.2.2 Fluorescence Microscopy Issues

Although very popular, fluorescence microscopy imaging (FMI) is not issue-free. Common issues we were faced with in FMI include:

- Non-linear illumination: this was a direct result of poor lighting conditions or sub-optimal experimental setup;

- Photobleaching (also known as fading): the irreversible decomposition of the fluorescent molecules in the excited state because of their interaction with molecular oxygen before emission [12];

- Largely out-of-focus image regions: again, as a result of sub-optimal experimental setup or microscope failure;

- Low or varying contrast: due to both photobleaching and different experimental setups;

- Gaussian noise and chromatic aberrations;

- Cellular and parasitic overlapping: due to various overlapping focal planes. This was, by far, the most challenging issue as we dedicated most of our efforts addressing it (sections 4.6 and 4.6);

# Chapter III

# Related Work

In this chapter we provide both the current state of the art in cellular and molecular imaging and the relevant background information needed for the understanding of the developed work. Regarding the state of the art we discuss some of the latest approaches in the fields that make up the typical pipeline for this type of problems (pre-processing, segmentation, machine learning algorithms and statistical methods such as mixture models). In the final section of this chapter we present a comprehensive analysis of the methods we deemed most fitting for this particular problem, focusing on the trade-offs between them.

## 3.1    State of the Art

### 3.1.1      Segmentation

The classic goal of segmenting an image is usually to partition said image into sets consisting of connected (by some metric) pixels.  This aims to simplify or change the representation of the image into a form that is easier to analyse [13]. In other words, image segmentation is a process by which each pixel is labelled, such that pixels with identical labels are considered part of the same region, through the fact that they share some common visual characteristic.

Segmentation techniques range from simple fixed threshold values, histogram-based adaptive thresholds and clustering techniques to partial differential equation-based methods, such as level-sets. An overview of the considered techniques for this work will be given in Section 2.2 - Background.

One of the first relevant approaches is the one described in [4], where cell clusters are detected in LUV colour space and delineated through their borders, by employing a gradient ascent mean shift method. In [5] segmentation is performed through a Canny edge detector, followed by a circle detection algorithm is proposed.

Another interesting approach in [6] proposes a thresholding scheme using information regarding the image's chromatic properties. This last approach inspired us to look for similar patterns in *Leishmania* images, which later then led to the first part of our segmentation step. Finally, a method for blood cell segmentation, which resorts to the EM algorithm to model cells as 3D Gaussian distributions, according to a feature vector composed of its pixel's HSV values is introduced in [7].

We would like to point out that the works in [5 and 7], although sound and requiring no interaction from the user, lack the ability to deal with clustered cells (and non-circular cells in [5] or Gaussian distributed regions in [7]).

## 3.1.2    Gaussian Mixture Models

A mixture model is a very powerful tool to statistically describe data. In our work's context it is generally used as a probabilistic model for the identification of sub-data-sets with a larger dataset, without requiring that the individual observations have any information as to which sub-data-set they belong. In other words, mixture models make statistical inferences about the properties of the sub-data-sets given only observations of the pooled data-set by estimating the most likely parameters for each sub-data-set (i.e.: the combination of parameters that best explains the observed data).

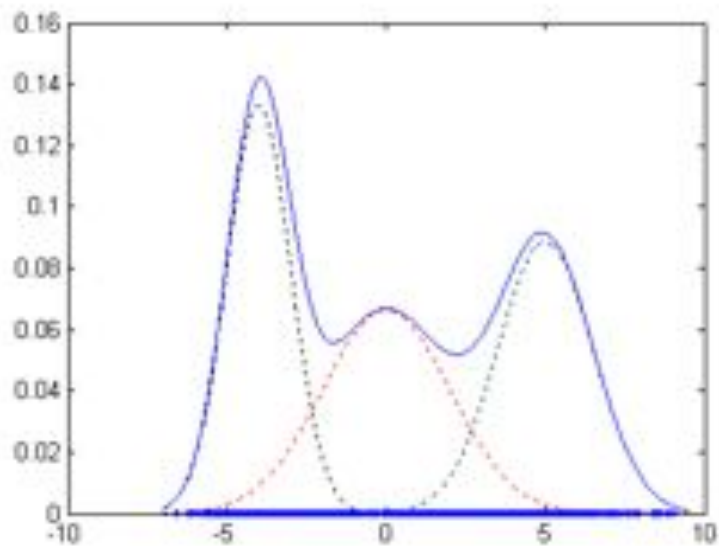

Figure 3.1.2:1 - Overall population (blue) and three modelled Gaussian sub-populations (black, red and green)[6].

Gaussian mixtures models (GMM) are mixture models that assume the observed data-set is composed of sub-data-sets with a Gaussian distribution and are one of the most popular forms of mixture models, mainly due to its real world applicability. One of the most popular methods for the implementation of a GMM is the Expectation-Maximization (EM) algorithm, as described by Bishop in [14].

GMMs usually need *a priori* information on the number of mixtures present in the overall dataset. One method to overcome this is cross-validation [15], which is a technique to estimate how accurately a predictive model will perform by dividing the original data set into *folds* (i.e.: partitioning it into various sub-sets of the same size). One *fold* is used to evaluate the built classifier/ model and the remaining *folds* to train/build it. While this method can increase the accuracy of said method by enhancing the chosen parameters it is inherently limited by how well we can evaluate the method's results.

This is especially true in GMMs, since we usually test the fit between the built model and the observed dataset with a log-likelihood ratio test [16]. It is therefore not difficult to conclude that the more mixtures we assume are present in the overall dataset, the higher the log-likelihood ratio test will score. This is, of course, because with more parameters available, the easier it will be for the algorithm to explain the observed data, especially in the presence of noise, missing data or extremely deformed Gaussian distributions. This issue will be addressed in more detail in section 4.5.2 and 4.6 of this thesis.

In face of these circumstances, there is a need to obtain robust parameter estimator functions. These are discussed further in the thesis, as we eventually encountered this problem. Some possible approaches are; binding the parameter interval through other contextual information; or using *a priori* information on the behaviour of the log-likelihood function to estimate optimum parameter interval, possibly coupling this method with the previous one. In [17], Zhuang has proposed a method similar to the aforementioned one by building an estimator for the number of Gaussian mixtures present in a dataset through the modelling of the log-likelihood function. While this method is able to sustain high variances in the Gaussian parameters and is robust to noise, the estimator takes a considerable amount of time to produce an output, which makes its usage in an online system unviable.

Gaussian Mixture Models have been used in a multitude of fields, ranging from finances and economics, to medical imaging and biometric verification. As an example, Brigo and Mercurio successfully used them in the development of financial return models [18 and 19] and Alexander applied them to market behaviour estimation [20].

In the field of handwriting recognition we point out the example given by Bishop in [16], where digit recognition is performed by building ten statistical models (one of each number from 0 to 9) and matching the statistical parameters of a new image to each of the pre-computed models.

With respect to speaker identification, Reynolds has studied how to obtain robust text-independent speaker identification using Gaussian mixture speaker models [21 and 22]. Stylianou has also used GMMs in multimodal biometric verification [23].

GMMs have also been applied to DNA protein sequence matching [24]. The algorithm described in this paper discovers one or more motifs in a collection of DNA or protein sequences through the EM algorithm to fit a two-component finite mixture model to the set of DNA sequences.

In [25] a method for real-time tracking of objects using adaptive background mixture models, which can be adapted to microscopy imaging is presented. Stauffer determines the background model by considering each pixel as a mixture of Gaussians and considering the persistence and variance of each of the Gaussian distributions in the image, determine which Gaussians may correspond to background colours.

### 3.1.3    Machine Learning

Machine Learning (ML) is a branch of artificial intelligence (AI) and, as a discipline, is concerned with the design and development of algorithms that allow computers to learn from the usage of empirical data (the repeated usage of this data is often referred to as experience). A widely quoted definition of the field is "A computer program is said to learn from experience E with respect to some class of tasks T and performance P, if its performance at tasks T, as measured by P, improves with experience E." [26].

ML algorithms can be organized into the following classes:

- Supervised learning: the algorithm is provided a training set consisting of labelled observations and produces an inferred function that should map any new input value to the correct output result based on what it *learned* from the training set;
- Unsupervised learning: the algorithm tries to find coherent sub-spaces within the dataset. Examples of unsupervised learning include, for example, clustering methods;
- Semi-supervised learning: is an hybrid approach between supervised and unsupervised learning that combines labelled and unlabelled observations;
- Reinforcement learning: the algorithm learns from the observation of its action on the world. Each action has a perceived impact in the world, meaning each action generates some form of feedback, which the algorithm uses to improve its odds of success;

- Transduction: consists of an approach that aims at not solving the more general problem simply by inducting sets of rules, but producing a more specific solution for the problem at hand. This is done by not using only the labelled data in the training set, but also the correlations between the unlabelled data to produce a solution for the specific problem at hand;
- Learning to learn: the algorithm estimates its own inductive bias based on previous experience;

Machine learning has been successful in a wide range of applications. It has been widely used for content-based classification of web pages such as the work presented by Esposito and Liu in [27 and 28], respectively. It has also found popularity in pattern recognition [29 and 30] and speech recognition software, such as:

- Dragon Naturally Speaking [31];
- Google Voice Search [32];

In the field of computer vision, Kubat has performed oil spill detection in satellite radar images [33], Sebastiani has automated text categorization [34] and Mitchell describes the usage of a massive-scale ML system for the automatic online recognition of letters with handwritten addresses in the US postal offices [35]. Finally, in the field of biology and medicine, several books on ML have been written. Some notable examples are the works of Goldberg, Baldi, Kononenko and Bishop in [36, 37, 38 and 14], respectively.

## 3.2 Cellular Microscopy Imaging

In this sub-section we discuss the most promising work directly related to our problem. Most of the referred work stops at the segmentation step, not trying to further segment regions with clustered cells or classifying them.

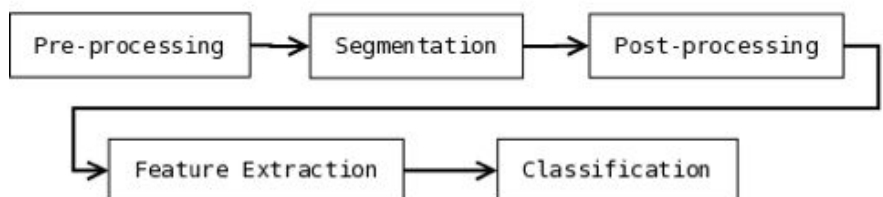

Figure 3.2:1 – Typical workflow found in the described algorithms. Our contribution is centred in the expansion of the classification step, by admitting the possibility of macrophages clusters generated by an imperfect segmentation. These are identified in the augmented classification step and further segmented into individual regions.

Liao et al. [39] use a simple threshold method, coupled with mathematical morphology and contextual information-based shape detection to detect white blood cells. However, this approach is sensitive to noise and irregular/non-uniform lighting and contrast conditions. While this is easily corrected with most of the standard methods for image pre-processing, its biggest drawback is that it does not tolerate cells outside the defined conditions (e.g.: poorly segmented regions forming a cell cluster region).

A fully automated membrane segmentation technique for immunohistochemical tissue images with membrane staining is described in [40]. This approach segments stained cellular membranes and allows the reconstruction of unstained tracts by using the nuclear membranes as a spatial reference. A white blood cell segmentation scheme using scale-space filtering and watershed clustering in HSV colour space is also proposed in [41].

Park explores bone marrow cell segmentation through a watershed algorithm that produces an over-segmented image that is then iteratively relaxed and classified as background, cytoplasm, cell nucleus or red blood cell according using fuzzy logic [42]. This work is sensitive to illumination and noise conditions, since it seems to overly rely on the mean colour values of each patch for the relaxation procedure. Begelman also performs cell nuclei segmentation using fuzzy logic [43]. Colour and shape features are used to detect cells, through a fuzzy logic engine. This work is more robust than the previous one because of the extracted shape features, which serve as an auxiliary classification input. However, it is still not able to account for non-circular or abnormally coloured cells due to the rule's simplicity.

In [44] Yu uses level sets in conjunction with the cell nuclei as seed points (obtained from a previous segmentation step) to detect the cytoplasmic boundaries, including irregularly shaped and overlapping ones. A similar approach to cell segmentation is proposed by Yan in [45]. The main difference between these two approaches is that Yan uses a distance map of the initial adaptive histogram-based thresholding step, which is then used to create a watershed transform that provides a region list that acts as the level-set seed points. The only drawback to this approach is that, once again, it is not able to deal with highly cluttered images, because the distance map would not provide enough information to accurately parameterize the watershed transform, thus leading to a wrong number/location of seed points for the level-set step.

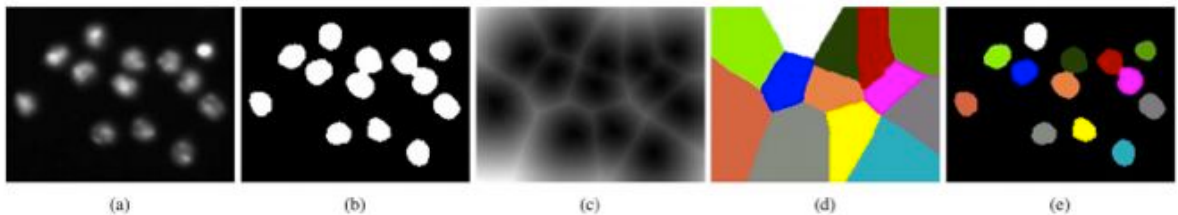

Figure 3.2:2 – Segmentation example of the method proposed in [45]. (a) Patch taken from original DNA image. (b) Binary thresholding result of (a). (c) Distance transform of (b); (d) Result of the watershed algorithm on (c). (e) Labelling nuclei by combining (b) and (d).

Although not directly related to our particular problem, there is much work done in microscopy imaging with video processing techniques. The papers presented in the next paragraph focus their efforts on cell detection and posterior declustering through frame correlation methods.

In [46] the use of time as an extra dimension as a means for determining cell trajectories from 2D + time data, through the use of Fast Marching level set methods [47] is studied. In [48] Zhou describes a hybrid segmentation approach using an adaptive threshold, mathematical morphology and a watershed transform. In his work, Zhou classifies cells accordingly to their phase, through a Markov Model [49]. Wang et al. also propose a similar method to Zhou's, by implementing an adaptive threshold, combined with a gradient vector field to detect cell nuclei in various time-lapse images, which are then used for a spatial correlation analysis [50].

In [51] the authors have devised an automatic marker-controlled watershed segmentation and mean-shift cell tracking for time-lapse microscopy. In their approach fine and coarse structures are used on a histogram thresholded image to obtain the initial markers for the watershed transform. The obtained regions are then tracked by employing an algorithm based on mean shift and a Kalman filter. This methodology has the advantage of being able to improve from frame to frame, since information on the cell number, position and markers can be used in subsequent frames.

## 3.3   Background

In this section we describe the methods and techniques used throughout the implementation period of this thesis. We also provide a quick overview of some discarded methods, followed by an analysis of why each method was (or not) chosen.

### 3.3.1   Pre-processing

On most occasions, image acquisition conditions are not under direct control, resulting in low SNR images. The aim of pre-processing techniques is to improve the overall image quality by improving contrast, lighting conditions and removing as much noise as possible.

**Histogram Equalization**

Histogram equalization is a common technique for enhancing the appearance of images, by widening peaks and condensing valleys in the image histogram. This tends to normalize illumination conditions, leading to more robust results for other computer vision algorithms such as, for example, segmentation ones. A non-linear/monotonic transfer function maps input to output pixel intensity values, in order to achieve this. In a discrete context, this can be described in the following way. Consider a discrete greyscale image {x} and let $n_i$ be the number of occurrences of the value $i$. The probability of randomly choosing a pixel with value $i$ in the image is:

$$p_x(i) = p(x{=}i) = n_i\,/\,n, \quad 0{<}i{<}L \qquad (1)$$

$L$ being the number of grey values present in the image, $n$ being the total number of pixels in the image, and $p_x(i)$ being the image's histogram for pixel value i, normalized to [0,1]. The cumulative distribution function, corresponding to $p_x$ is:

$$cdf_x(i){=} \Sigma^{i}_{j=0} p_x(j), \qquad (2)$$

, which is also the image's accumulated normalized histogram.

We want to create a monotonic transformation of the form $y = T(x)$ to produce a new image {$y$}, such that its CDF will be linearized across the value range, i.e.

$$cdf_y(i) = iK \qquad (3)$$

for some constant $K$. This transformation can be written as:

$$y = T(x) = cdf_x(x) \qquad (4)$$

**Contrast Stretching**

Contrast stretching (sometimes known as normalization, or dynamic range extension in the field of digital signal processing) is a technique meant for improving an image by stretching the range of intensity values contained in it, enabling it to use its full dynamic range.

To perform a contrast stretch, the first step is to determine the maximum and minimum intensity values over which the image will be extended. These will constitute the lower and upper bounds, henceforth known as a and b, respectively (in 8 bit images, as in our case, these values are 0 and 255). Next, we examine the original image histogram to determine the value limits in the unmodified picture. This range can then be stretched linearly with original values, which lie outside the range (i.e.: each original pixel value *r* is mapped to an output value *s*). This can be expressed in the following equation:

$$s = (r - c)\left(\frac{b - a}{d - c}\right) + a \tag{5}$$

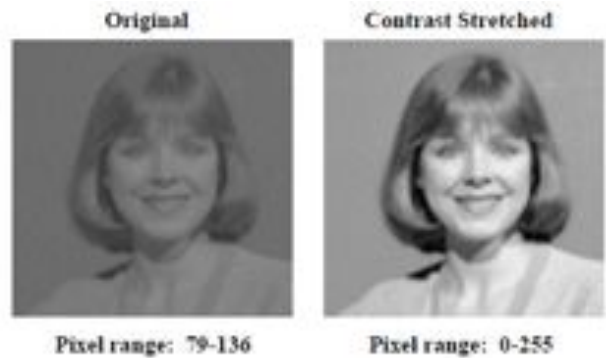

Figure 3.3.1:1 - Typical example of a contrast stretching operation[7].

Unlike other methods, such as the Histogram Equalization, this operation is a linear mapping of input to output values, which preserves the relative distances between luminosity values, rather than warping them (and losing information in the process).

**Gaussian filter**

A Gaussian filter is a filter whose impulse response is a Gaussian function. In two dimensions, as is the case in digital images, its impulse is given by the product of two Gaussians (one per direction):

$$g(x, y) = \frac{1}{2\pi\sigma^2}e^{-\frac{x^2+y^2}{2\sigma^2}} \tag{6}$$

---

7 *Adapted from: http://homepages.inf.ed.ac.uk/rbf/HIPR2/stretch.htm.*

It is normally used to remove detail or smooth images affected by additive Gaussian noise. Its main drawback is that while it eliminates fine texture and fine grain noise, it also dissipates borders and other high contrast features, so it is usually used with small kernels (3x3 or 5x5 in most cases).

**Mathematical Morphology**

Mathematical Morphology is a tool for extracting image components that are useful for representation and description. It is a set-theoretic method of image analysis providing a quantitative description of geometrical structures. Morphology can provide boundaries of objects, their skeletons, and their convex hulls. It is also useful for many pre- and post-processing techniques, especially in edge thinning and pruning.

Most morphological operations are based on simple expanding and shrinking operations. The primary application of morphology occurs in binary images, though it is also used on grey level images. The most common operations are *erosion* and *dilation*. Considering a target object $A$ and a structuring object $B$, a *dilation* can be defined as $A \oplus B = \{x : \hat{B}_x \cap A \neq \emptyset\}$. In a similar fashion, an *erosion* can be defined as $A \ominus B = \{x : B_x \subseteq A\}$.

These operations are easily understood and implemented as convolution masks, as illustrated in Figure 3.3.1:2.

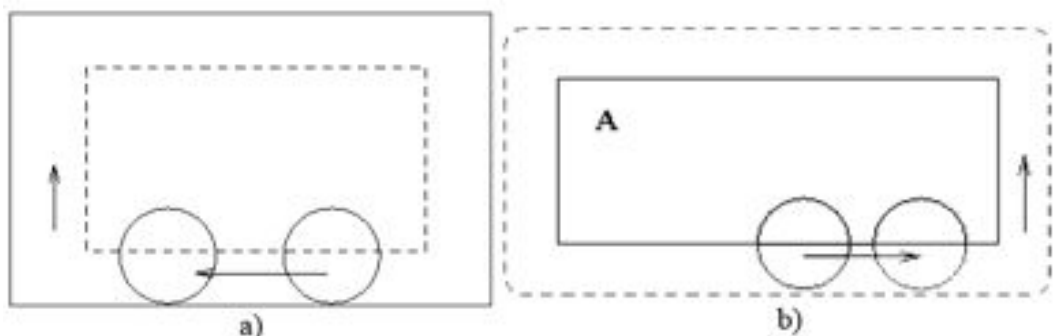

Figure 3.3.1:2 - Mathematical morphology operations[8]. a) Erosion; b) Dilation.

Other popular operations are the *opening* and *closing* operations, which are the consecutive application of an *erosion* followed by a *dilation* and a *dilation*, followed by an *erosion*, respectively.

---

### 3.3.2  Segmentation

#### Fixed Threshold

Thresholding is the simplest form of segmentation. It consists of dividing the image pixels in two sets (usually called background and foreground). This is achieved by a simple function $F$ that acts on each pixel $p$ with a single parameter $T$, called the threshold. A pixel is considered to belong to the background if $f(p) < T$ and belonging to the foreground if $f(p) \geq T$.

While this approach is quite simple, it lacks robustness as images are not always so well organized that a single value can be defined for whatever number of sets (classes) are present, much less can a fixed value be optimal for more than a few images.

#### Otsu's Method

Otsu's method is a histogram based thresholding technique used to automatically perform image segmentation. It was proposed by Otsu in his original paper [52] and has been one of the most successful simpler segmentation techniques to date. The algorithm assumes that the image to be thresholded contains two classes of pixels and computes the optimum threshold that separates these two classes so that their intra-class variance is minimal [53]. Otsu's method performs and exhaustive search for the minimizing threshold for the intra-class variance, or maximum inter-class variance:

$$\sigma_b^2(t) = \sigma^2 - \sigma_w^2(t) = \omega_1(t)\omega_2(t)\left[\mu_1(t) - \mu_2(t)\right]^2 \qquad (7)$$

This is defined as a weighted sum of variances of the two referred classes:

$$\sigma_w^2(t) = \omega_1(t)\sigma_1^2(t) + \omega_2(t)\sigma_2^2(t) \qquad (8)$$

Weights $\omega_i$ are the probabilities of the two classes separated by a threshold $t$ and $\sigma_i^2$ variances of the classes. These weights $\omega_i$ and the class means $\mu_i$ are updated after each iteration of the algorithm, until a maximum inter-class variance is found. This concept can easily be extended to multiple thresholds, when multiple classes are found in the image, resulting in a multi-level Otsu threshold.

The method is at a disadvantage when the image does not have a bimodal (or n-modal for a multi-level Otsu threshold) distribution. It will also perform under expectations when the classes are very unequal or in the presence of variable illumination.

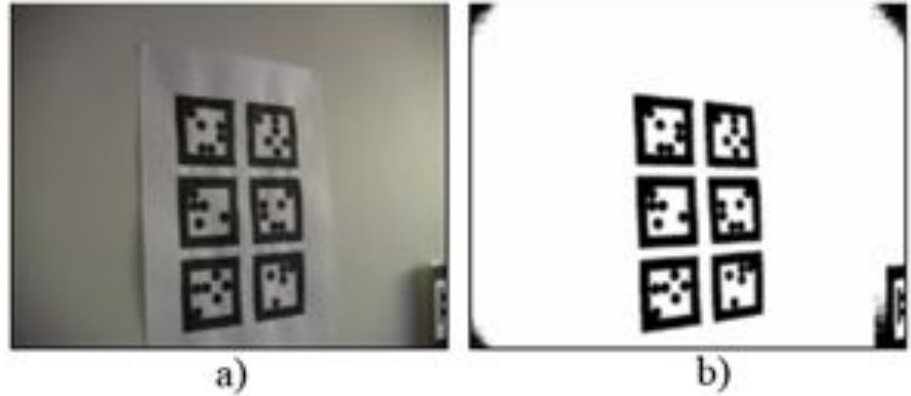

Figure 3.3.2:1- Otsu's Method example[9]. a) Original image under normal illumination. b) Thresholded image.

### K-means Segmentation

The K-means algorithm is a statistical method for cluster analysis, which aims to partition a set $X$ of $n$ observations into $k$ clusters, where each observation $s_i$ belongs to the cluster with the nearest mean value. Given $X$:($x_1, x_2, .. , x_n$), where each observation is a d-dimensional vector, k-means intends to map all $n$ observations into $k$ clusters, so as to minimize the following error function:

$$\arg\min_s \sum_{i=1}^{k} \sum_{x_j \in S_i} \left\| x_j - \mu_i \right\|^2 \qquad (9)$$

where $u_i$ is the mean of each cluster.

The most popular algorithm is the Lloyd's algorithm [54]. It uses an iterative refinement technique consisting of two steps:

Assignment step: Assign each observation to the cluster with the closest mean:

$$S_i^{(t)} = \left\{ \mathbf{x}_j : \left\| \mathbf{x}_j - \mathbf{m}_i^{(t)} \right\| \leq \left\| \mathbf{x}_j - \mathbf{m}_{i^*}^{(t)} \right\| \text{ for all } i^* = 1, \ldots, k \right\} \qquad (10)$$

Update step: Calculate the new means for each cluster centroid;

$$\mathbf{m}_i^{(t+1)} = \frac{1}{|S_i^{(t)}|} \sum_{\mathbf{x}_j \in S_i^{(t)}} \mathbf{x}_j \qquad (11)$$

---

9 *Adapted from: http://www.dandiggins.co.uk/arlib-3.html.*

The obtained clusters are then considered regions of interest, resulting in a segmented image. In some cases, if the image is sufficiently complex, they may be further analysed and grouped into several groups, each representing a distinct class.

This method's main limitations come from being a heuristic method, from its cluster model and from the number of clusters *k* being an input parameter. Since this method is a heuristic method there is no assurance that it will converge to a global optimum. Regarding it's cluster model, while the Euclidean distance makes the algorithm efficient, it also makes it biased to clusters of similar sizes, which often fails to correctly cluster datasets where this assumption does not hold true. Finally, having to *a priori* know the number of clusters can be quite troublesome (e.g.: ill-segmented cellular regions). Pelleg and Moore et al. have improved the K-means algorithm by introducing an efficient estimation mechanism, resulting in the X-means algorithm [55]. Even so, the cluster model remains the same and can be a serious drawback when the clusters are not uniform in size or shape.

### Mean Shift Segmentation

Mean shift is a non-parametric technique originally proposed by Fukunaga and Hostetler in [56] for analysing complex multimodal feature spaces and estimating the stationary points (modes) of the underlying density function. It does not require prior knowledge on the cluster number and does not constrain the cluster shape or size, as it is based on a kernel density estimator.

Mean shift treats the points on the feature space as a probability density function. Dense regions in the feature space correspond to local maxima or modes. So, for each data point, a gradient ascent is performed on the local estimated density, until convergence. The stationary points obtained via gradient ascent represent the modes of the density function. All points associated with the same stationary point belong to the same cluster. The mean shift is called *m(x)* and so, we can summarize the procedure as:

```
for each point xᵢ:
   compute the mean shift vector m(xᵗⱼ)
   move the density estimation window by m(xᵗⱼ)
   repeat until convergence
return
```

The main drawbacks reside in the kernel choice. Choosing an inappropriate kernel can lead to under (or over) segmentation and larger kernel values will considerably increase the algorithm's runtime making it somewhat cumbersome for higher resolution images and useless for real-time applications.

### Watershed Transform

The watershed transform is a mathematical morphology based-segmentation technique proposed by Digabel and Lantuéjoul [57] that obtains its inspiration from geography. There are two popular analogies for this method, the drainage and the immersion analogy. In the drainage analogy we consider the image as a topological relief, with the gradient function providing the height value for each pixel and simulate rainfall in the topographical map. Naturally, water will flow in the path of steepest descent and accumulate in the lowest point that particular path leads, immersing the image. In the immersion analogy, the idea is to, again, imagine the image as a topographical map, but this time being immersed with water, not from above but from holes in the local minima. In time, water will fill up the image, starting with these local minima, eventually forming catchment basins and merging regions where water meets.

This technique usually results in an over-segmented image, so various methods have been design to counter this. Two basic but popular methods consist in removing irrelevant minima by modifying the gradient function used or using a sub-region merging step. The most used method is to first automatically detect markers that define the immersion points and minimum water levels. As there are several definitions and formalizations for the watershed technique, we refer the reader to the survey by Roerdink and Meijster [58], for a very complete overview of this method.

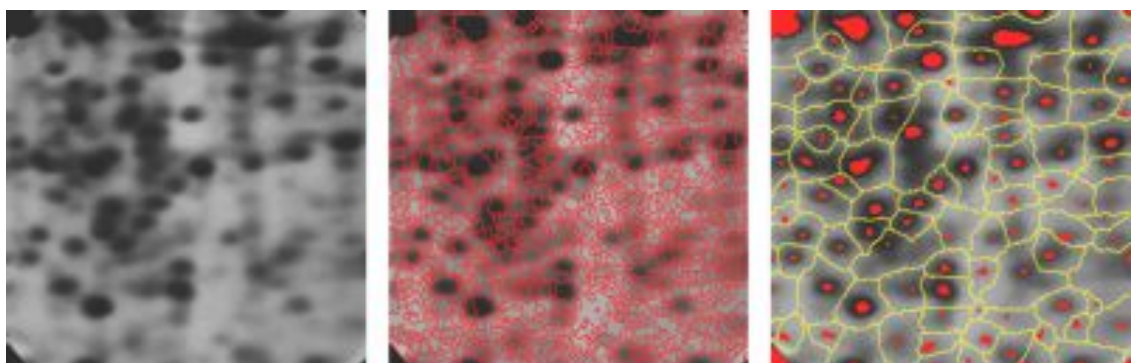

Figure 3.3.2:2- Watershed transform example[10]. From left to right: original image, over-segmented watershed transform, marker (red) based-watershed transform.

### Connected Component Analysis

Although not a segmentation technique per se, Connected Component Analysis (CCA), or Connected Component Labelling (CCL) is an algorithm based on graph theory for connecting subsets of related pixels in an image, i.e. all pixels in a connected component share some similar characteristic

---

*10 Adapted from Image Segmentation and Mathematical Morphology, © Serge Beucher, 2010.*
*http://cmm.ensmp.fr/~beucher/wtshed.html.*

(normally an intensity value or interval) and are in some way connected with each other. In most cases this technique is used to obtain the regions present in a binarized image (as the binarization process may have only classified each individual pixel). Pixels are usually also considered connected according to a 4 or eight neighbour scheme, as show in Figure 3.3.2:3.

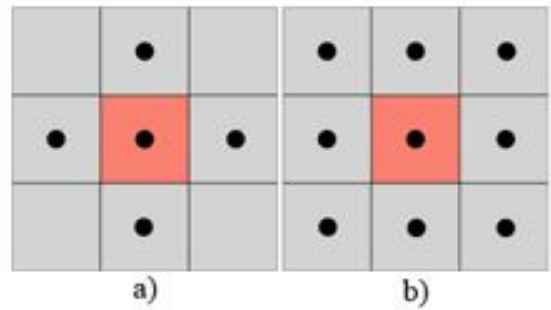

Figure 3.3.2:3 - Pixel connectivity schemes. a) 4-neighbor connectivity. b): 8-neighbor connectivity.

The most popular algorithm for this task is the two-pass algorithm, which iterates through 2-dimensional binary data. The first pass intends to record equivalences and assigns temporary labels to each pixel, while the second pass refines the result of the first pass by merging connected pixels with different labels. For more details please refer to the latest work by Gonzales and Woods, found in [59].

### 3.3.3  Feature Extraction

**Shape descriptors and Chain codes**

Shape descriptors are commonly used to describe two-dimensional shapes. Shape descriptors can be divided into three main categories:

1) Contour based descriptors: the contour of the object is mapped to some representation from which a shape descriptor is derived (Fig 11.a));

2) Image based descriptors: the shape descriptor is built using features computed by analysing the description of the region occupied by the object on the image plane;

3) *Skeleton based descriptors:* after a skeleton is computed, it is mapped to a tree structure that forms the shape descriptor;

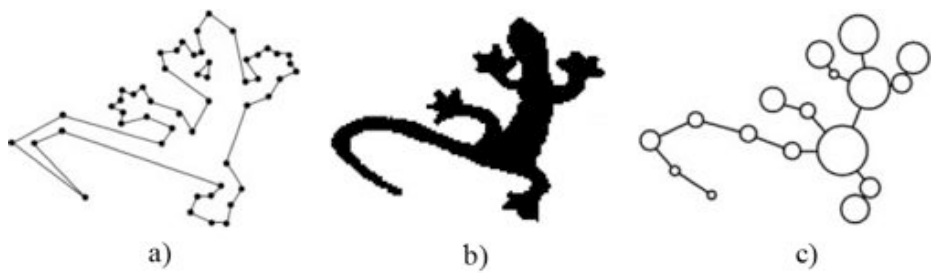

Figure 3.3.3:1 - Various shape descriptors for an object[11]. From left to right: contour-based descriptor, image-based descriptor and skeleton-based descriptor.

There are various desirable properties shape descriptors should have for object recognition:

1) Uniqueness: each object should have an unique representation;

2) Completeness: there should be no ambiguous representations;

3) Invariance under geometrical transformations: they should be invariant under translation, rotation, scaling and reflection operations;

4) Sensitivity: they should be able to easily represent differences between similar objects;

5) Abstraction from detail: they should be able to represent the basic features of a shape, while being able to abstract from detail.

A very large variety of shape descriptors exist [60] and we will only describe a few simple ones that were considered in the scope of this thesis.

**Chain Codes**

Chain Codes are lossless compression algorithms, particularly effective in images made up of large connected objects. The principal behind these codes is to separately encode each object by creating a list, which reflects the behaviour of the object's boundary. The chain is created by moving an encoder along the object's boundary and, for each position, transmitting a symbol that represents the direction of the current boundary movement. This process is repeated until the encoder reaches the starting point. The Freeman Chain Code [61] is the most popular and well-known example.

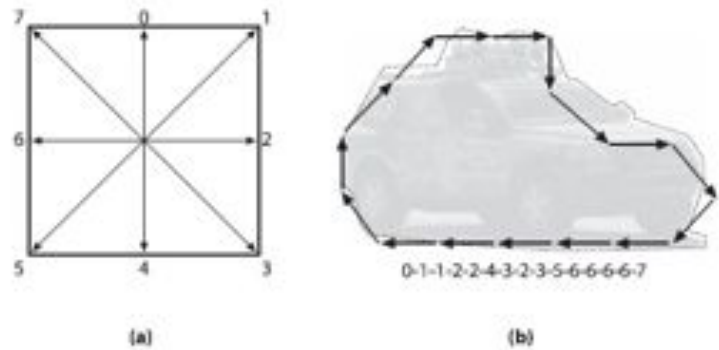

Figure 3.3.3:2 – Object description through a 8-connected Freeman Chain Code[12]. a) Representation of an 8-connected scheme. b) Object description by boundary direction analysis (over a sliding window).

### Centre of Mass

The centre of mass of an object is the mean location of all the mass in a system. It can be defined as the average of the positions of each particle in the object, weighted by their masses:

$$\mathbf{R} = \frac{\sum m_i \mathbf{r}_i}{\sum m_i}. \tag{12}$$

Where, $R$ is the centre of mass, $m_i$ the individual mass of each particle and $r_i$ their respective weights. In our case we can either consider that each particle has the same weight or that it's weight is given by the pixels intensity or gradient value. Although the centre of mass is influenced by the object's shape, thus providing us with some information it can be ambiguous, as many different shapes may produce the same centre of mass.

### Minimum Bounding Box

A minimum bounding box (or minimum bounding rectangle) is the smallest enclosing box for a set of points in an N-dimensional space. They are usually used as a (sometimes rough) indication of an object's size and relative proportions.

---

*12 Adapted from: Learning OpenCV: Freeman Chain Codes, Polygon Approximations and Dominant Point. URL:*

*http://sapachan.blogspot.com/2010/04/learning-opencv-freeman-chain-codes.html.*

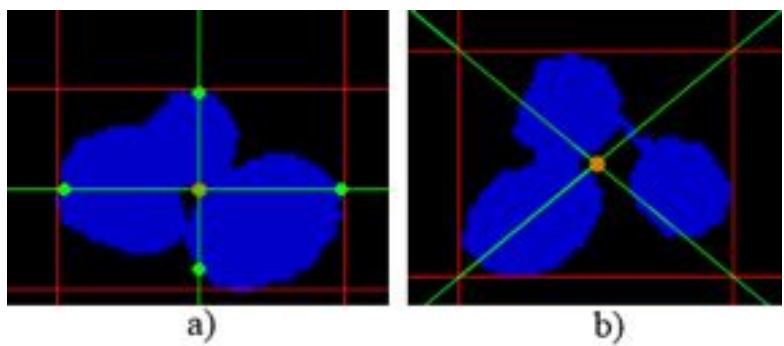

Figure 3.3.3:3 – Minimum bounding box examples (in red).

## 3.3.4 Classifiers

Classifiers are methods used to map an observation into a set of classes. They consist in mapping functions that given an observation (usually denoted by a feature vector $F$ with $n$ features, each with $k$ coefficients) output the class that observation belongs to.

Classifiers can process both labelled and unlabelled training sets (i.e. the training data has, or not, information on what class each observation belongs to). The general approach is, however, quite different. When the training set is labelled it is known as supervised classification and when it is unlabelled it is known as unsupervised classification (or clustering).

Supervised classifiers can be built using a large number of methods. Among these the simplest one is to build a set of rules that define how to separate the feature space and, therefore, to classify every observation. This method assumes *a priori* knowledge of the classification problem and can turn out to be too rigid if the model defined by the rules is too simple. Other methods, such as nearest neighbours use some metric (such as the Euclidean distance, class mean or Mahalanobis distance) to estimate the nearest neighbour to each new observation in the feature space [62].

Unsupervised classifiers try to find a hidden structure in the training dataset. Popular approaches include the use of density estimation techniques, such as mixture models, k-means or hierarchical clustering. Neural networks have also proven a quite popular means to build these classifiers [62].

### 3.3.5  Pattern Recognition

Duda and Hart defined pattern recognition (PR) as a field concerned with machine recognition of meaning regularities in noisy or complex environments [63]. A typical pattern recognition system consists of a sensor, which gathers a set of observations to be classified; a feature extraction mechanism that computes numeric (or symbolic) information from the observations; and a classification method that does the actual process of classifying the aforementioned observations, by relying on the extracted features.

The classification scheme is usually based on the availability of a set of patterns that have already been classified (widely referred to as the training set and its data as labeled). When labeled data is used, the resulting learning strategy is characterized as supervised. Should the classification scheme not be given an *a priori* labeled dataset, the learning strategy is classified as unsupervised. In this type of learning strategy, the method tries to establish the classes by itself, by searching for statistical regularities on the given data.

Classification methods usually use one of the following approaches: statistical, syntactic, or neural. Statistical pattern recognition (e.g. Gaussian Mixture Models) is based on statistical characterizations of patterns, assuming these are generated by a probabilistic system. Structural pattern (e.g. decision trees, support vector machines) recognition is based on the structural interrelationships of features and neural pattern recognition (feed-forward neural networks) employs the neural computing paradigm that has emerged with neural networks.

### 3.3.5.1 Supervised Methods

#### Artificial Neural Networks

An Artificial Neural Network (ANN) is a mathematical/computational model that attempts to simulate the structure of biological neural systems. They consist of an interconnected group of artificial neurons that process information using a connectionist approach. In most cases an ANN is an adaptive system that changes its structure based on external or internal information that flows through the network during the learning phase.

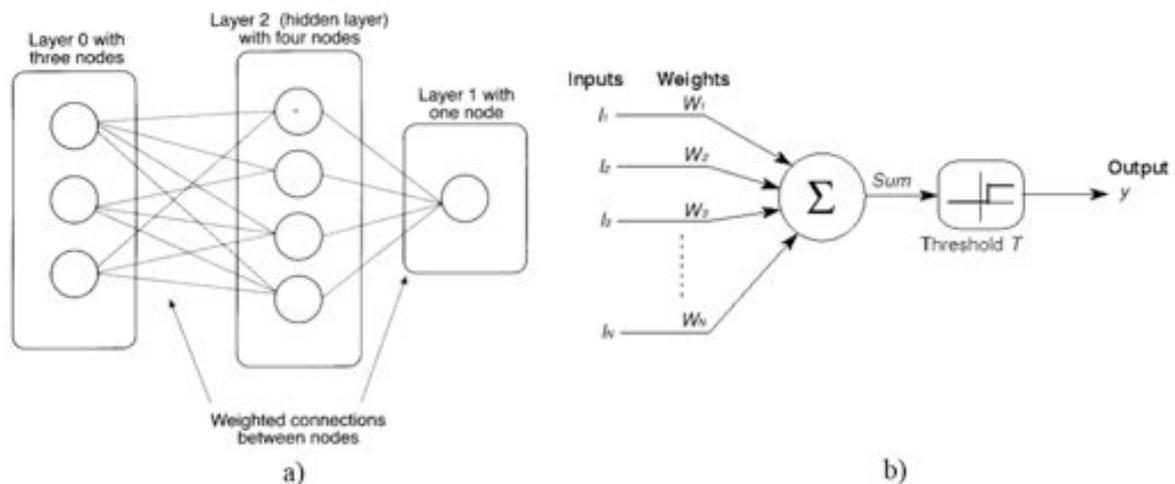

Figure 3.3.5:1 – Feed-forward neural network (FFNN) example[13]. a) FFNN architecture. b) McCullogh-Pitts neuron model.

Neurons are identical units (identified by circles in the picture above) that are connected by links (synapses that extend from a dendrite to an axon). These synapses are used to send signals from one neuron to the other [64]. Each link has a weight that defines its importance to the model. To create a model using an ANN, the link weights must be adjusted using the training data and a training algorithm.

ANN present a number of advantages, such as:

- They require a less formal statistical training than other methods;
- They are able to implicitly detect complex nonlinear relationships between dependent and independent variables;
- They are (also) able to detect interactions between predictor variables;
- Multiple training algorithms are available;

Amongst the disadvantages are:

- Their "black box" nature;
- A greater computational burden (as compared to the other methods referred);
- Proneness to over-fitting;
- Their model development's empirical nature;

**Decision Trees**

Decision Trees use a tree data structure for decisions with their outcomes at each leaf node. Figure 3.3.5:2 represents a classical decision tree example to check if the weather is good enough to play golf.

---

13 *Adapted from: http://wwwold.ece.utep.edu/research/webfuzzy/docs/kk-thesis/kk-thesis-html/node8.html.*

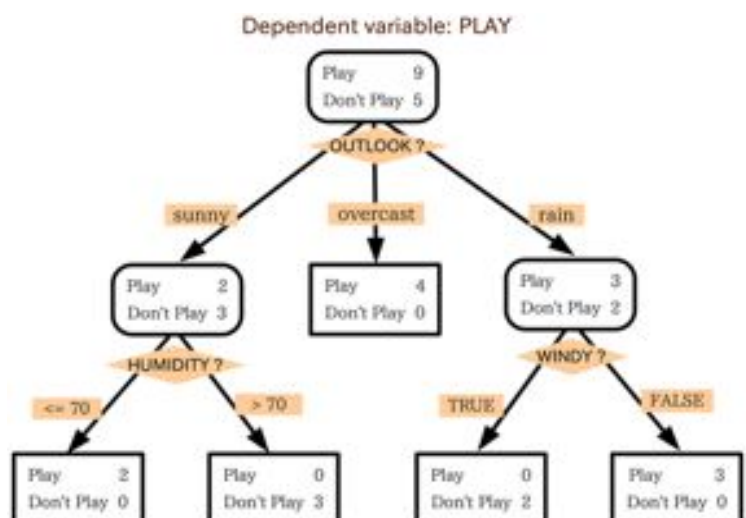

Figure 3.3.5:2 - Classical decision tree example[14].

Decision trees have various advantages over other classifiers, such as:

- They are simple to understand and interpret, both conceptually and visually;
- Little data preparation is required;
- Both numerical and categorical data are supported;
- They use a white box model;
- It is possible to validate a model using statistical tests;
- Very efficient even when the training data is large.

Some important drawbacks consist of:

- Creating an optimal decision tree is a NP-complete problem, which means that the decision tree creation algorithms won't always generate the best solution;
- There is a possibility the output model will over fit, i.e. the tree may not generalize the data well.

### Support Vector Machines

Support vector machines (SVM) are a technique based on statistical learning theory, which works very well with high-dimensional data reducing the curse of dimensionality problem [65]. The objective is to find the optimal separating hyper plane between two classes by maximizing the margin between the classes' closest points. This is done with the aid of support vectors, which are influenced by each observation in the dataset, as observed in Figure 3.3.5:3.

---

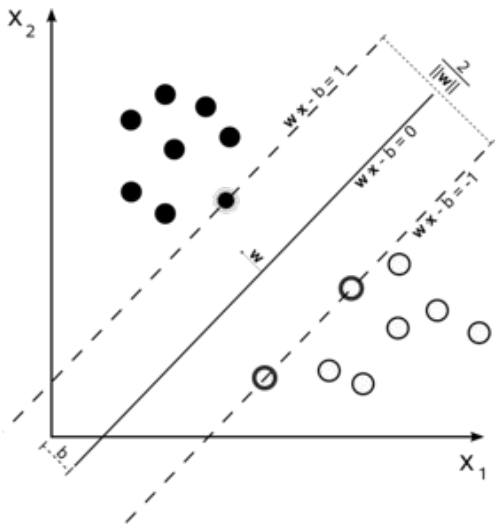

Figure 3.3.5:3 - Maximum-margin hyper-plane and margins for a SVM[15].

SVMs do, however, have another great asset. They are able to solve problems other methods could not possibly classify in the feature space by introducing them to a higher dimension plane. Consider the following figure:

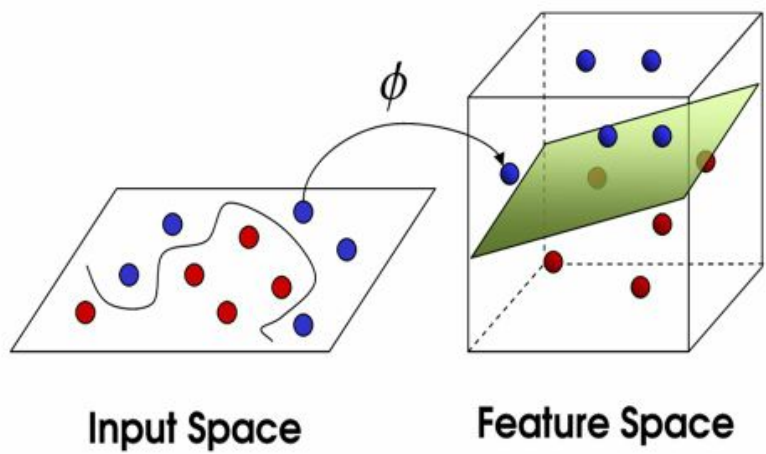

Figure 3.3.5:4 – Visual representation of solving an untreatable problem by introducing a higher dimension, through a kernel transformation ($\phi$)[16].

A classifier trying to classify the two classes present in the input space (red and blue dots, respectively) would either not be able to, or do a poor job at it. It would do so because it would have to draw a straight line in the input space, effectively including some red (or blue) dots in the wrong class. SVMs can, unlike some other methods, apply a *"kernel trick"*, as proposed by Boser, Guyon and Vapnik in [66] (and originally by Aizerman in [67]), to transform the feature space into a higher dimensional one, where classes are separable by a hyperplane. This is represented in Figure 3.3.5:4 by the kernel transformation $\phi$ and the green plane.

Some kernel transformations include:

*15 Adapted from: http://reference.findtarget.com/search/Support%20vector%20machine/.*
*16 Adapted from: http://www.imtech.res.in/raghava/rbpred/algorithm.html.*

- Homogeneous Polynomial: $k(x_i, x_j) = (x_i . x_j)^d$
- Inhomogeneous Polynomial: $k(x_i, x_j) = (x_i . x_j + 1)^d$
- Gaussian Radial Basis Function: $k(x_i, x_j) = exp(-\gamma \|x_i - x_j\|^2)$, for $\gamma > 0$
- Hyperbolic Tangent: $k(x_i, x_j) = tanh(kx_i . x_j + c)$, for some $k > 0$ and $c < 0$

The main advantages of SVMs are [68]:

- They exhibit good generalization;
- Learning involves the optimization of a convex function, so no false minima exist, unlike in neural networks;
- Few parameters required for tuning the learning machine (unlike neural network where the architecture and various parameters must be found);
- They are able to implement confidence measures.

### 3.3.5.2 Unsupervised Probabilistic Methods

#### Gaussian Mixture Models

A Gaussian Mixture Model (GMM) is a probabilistic model that aims at identifying the sub-populations present within a larger population. This model does not require an already observed dataset and identifies the sub-population to which one individual observation belongs. In other words, mixture models are used to make statistical inferences about the properties of the sub-populations given only observations on the overall population, without sub-population-identity information. A GMM is a weighted sum of $M$ component Gaussian densities as given by the equation:

$$p(\mathbf{x}|\lambda) = \sum_{i=1}^{m} w_i \, g(\mathbf{x}|\boldsymbol{\mu}_i, \boldsymbol{\Sigma}_i),$$

*(13)*

where $x$ is a D-dimensional continuous-valued data vector, $w_1, w_2, .., w_M$ are the mixture weights, and $g(x|\mu_i, \Sigma_i)$, $i = 1, ..., m$ are the component Gaussian densities. Furthermore, the mixture weights must satisfy the constraint: $\sum_{i=1}^{m} w_i = 1$ [69]. Each component is a D-variate Gaussian function of the form:

$$g(\mathbf{x}|\boldsymbol{\mu}_i, \boldsymbol{\Sigma}_i) = \frac{1}{(2\pi)^{D/2}|\Sigma_i|^{1/2}} \exp\left\{-\frac{1}{2}(\mathbf{x} - \boldsymbol{\mu}_i)' \, \Sigma_i^{-1} \, (\mathbf{x} - \boldsymbol{\mu}_i)\right\},$$

*(14)*

with mean vector $\mu_i$ and covariance matrix $\Sigma_i$.

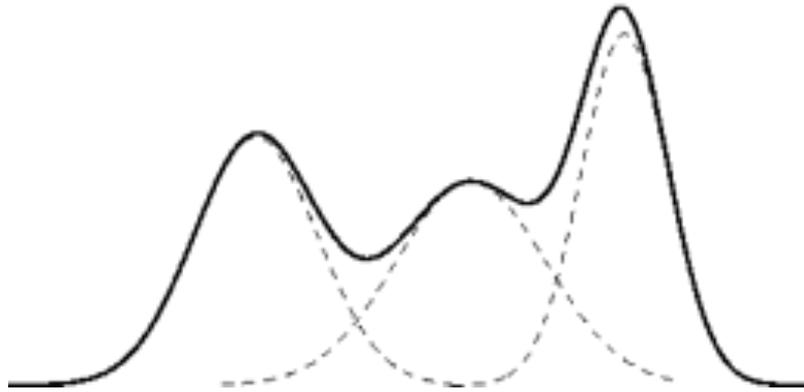

Figure 3.3.5:5 - Sub-set prediction of three Gaussian mixtures (dotted lines) in a larger dataset (bold line) through the use of GMMs[17].

Extracting information from an unlabelled dataset can only be possible if certain assumptions are made [63]:

- That the samples come from a known number of classes ($\lambda$);

- The *a priori* probabilities for each class $w_i$ are known;

- The forms of the class-conditional probability densities are known for all classes, $i = 1, .. , \lambda$ (which we assume to be the sum of M multivariate Gaussian probability density functions);

- The unknowns are the values of the parameter vectors (the respective weights of the M Gaussian probability density functions, as well as their mean value and covariance matrix, for each class);

### Expectation Maximization Algorithm

One of the most popular ways to implement GMMs is by employing a maximum likelihood algorithm to estimate the unknowns of the aforementioned statistical model [70]. In this thesis we used the Expectation Maximization algorithm, proposed in [71].

The EM (Expectation-Maximization) algorithm estimates the parameters of the multivariate probability density functions in the form of a Gaussian mixture distribution with a specified number of mixtures. Consider the following set of feature vectors $x_1, x_2, ..., x_N$ : N vectors from a d-dimensional Euclidean space drawn from a Gaussian mixture:

$$p(x; \theta) = \sum_{k=1}^{m} \pi_k p_k(x), \quad \pi_k \geq 0, \quad \sum_{k=1}^{m} \pi_k = 1, \qquad (15)$$

$$p_k(x) = \varphi(x; a_k, S_k) = \frac{1}{(2\pi)^{d/2} \mid S_k \mid^{1/2}} exp\left\{ -\frac{1}{2}(x - a_k)^T S_k^{-1}(x - a_k) \right\}, \qquad (16)$$

---

17 Adapted from: http://courses.ee.sun.ac.za/Pattern_Recognition_813/lectures/lecture06/node1.html.

Where $m$ is the number of mixtures, $p_k$ is the normal distribution density with the mean $a_k$ and covariance matrix $S_k$, $\Pi_k$ is the weight of the $k^{th}$ mixture. Given the number of mixtures $m$ and the samples $x_i : i = 1..N$ the algorithm finds the maximum-likelihood estimates (MLE) of the all the mixture parameters, i.e. $a_k$, $S_k$ and $\Pi_k$.

$$L(\mathbf{X}, \theta) = log\, p(x, \theta) = \sum_{i=1}^{N} log\left(\sum_{k=1}^{m} \pi_k p_k(x)\right) \to \max_{\theta \in \Theta}, \qquad (17)$$

$$\Theta = \left\{ (a_k, S_k, \pi_k) : a_k \in \mathbb{R}^d, S_k = S_k^T > 0, S_k \in \mathbb{R}^{d \times d}, \pi_k \geq 0, \sum_{k=1}^{m} \pi_k = 1 \right\}. \qquad (18)$$

EM algorithm is an iterative procedure, where each iteration consists of two steps. In the first step (E-step), we find the probability $\alpha_{ki}$ of sample $i$ belonging to a mixture $k$, using the currently available mixture parameter estimates:

$$\alpha_{ki} = \frac{\pi_k \varphi(x; a_k, S_k)}{\sum\limits_{j=1}^{m} \pi_j \varphi(x; a_j, S_j)}. \qquad (19)$$

In the second step (M-step) the mixture parameter estimates are refined using the computed probabilities:

$$\pi_k = \frac{1}{N} \sum_{i=1}^{N} \alpha_{ki}, \quad a_k = \frac{\sum\limits_{i=1}^{N} \alpha_{ki} x_i}{\sum\limits_{i=1}^{N} \alpha_{ki}}, \quad S_k = \frac{\sum\limits_{i=1}^{N} \alpha_{ki}(x_i - a_k)(x_i - a_k)^T}{\sum\limits_{i=1}^{N} \alpha_{ki}}, \qquad (20)$$

## 3.4 Software

Various tools for data mining and cell imaging analysis are available. In this section we overview some of the most relevant. We start by introducing two of the most popular machine learning softwares (WEKA and Rapid Miner). These provide the user with implementations of some of the aforementioned algorithms, while also accelerating and easing the creation of classifiers based on said algorithms. We the present a brief overview of the most popular image processing programs, available for the creation of automatic analysis routines in cellular and molecular images. We highlight their main features and available methods for the creation of said routines.

### Weka

The Waikato Environment for Knowledge Analysis (WEKA) [72] is one of the oldest and most used software for data mining. Weka features implementations of the most important open source data mining algorithms and is easy to integrate with custom software. This is because it provides a complete and easy to use JAVA API. Weka also comes with the Weka Explorer, which is a graphics user interface (GUI) that allows to easy usage of all Weka functionalities in a *standalone* fashion.

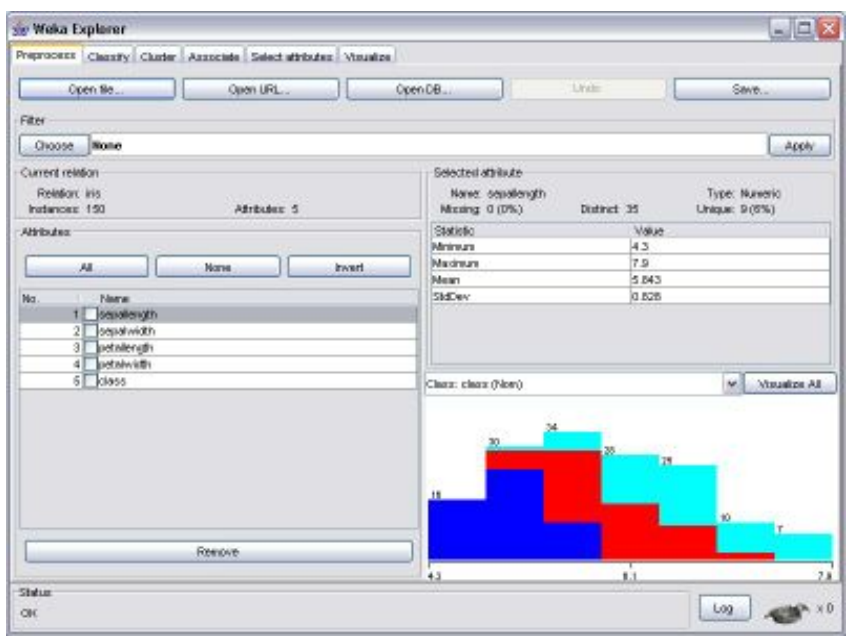

Figure 3.4:1 - Weka Explorer GUI[18].

**Rapid Miner**

Rapid Miner [73] is an open-source system for data mining. It features a GUI for the design of analysis procedures, which allows modularity and breakpoints. Other features include a high number of operators as well as many data transformations. In its recent updates Rapid Miner has been tuned for speed and is now able to process large amounts of data on the fly. Its scalability has seen an increase as a direct consequence of these improvements.

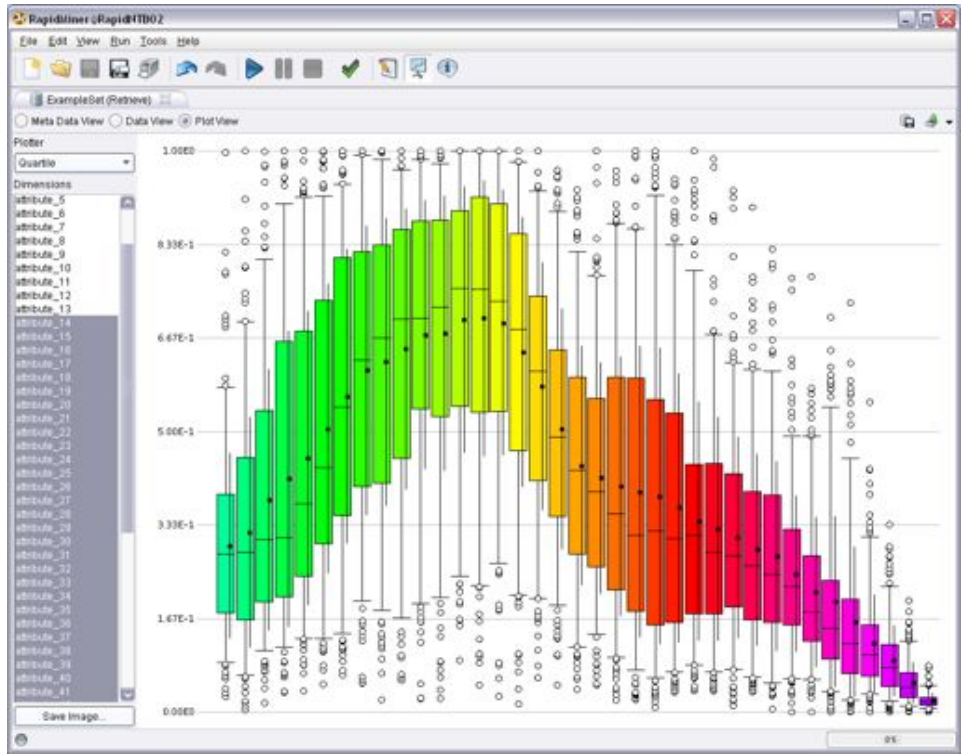

Figure 3.4:2 - Rapid Miner GUI[19].

**ImageJ**

ImageJ [74] is an open-source, Java-based image-processing program. It was designed with an open architecture and provides extensibility through Java plugins and recordable macros. Because it is based in Java, it is multi-platform and runs in both 32 and 64-bit modes. Currently, it is able to display, edit, analyse, process and save the most common image formats (including raw). Although it is based on Java, it's quote as being able to filter high-resolution images in under a tenth of a second. It also supports most of the typical computer vision algorithms, such as smoothing, contrast stretching, regions of interest (ROI) and edge detection, as well as manual image edition tools.

---

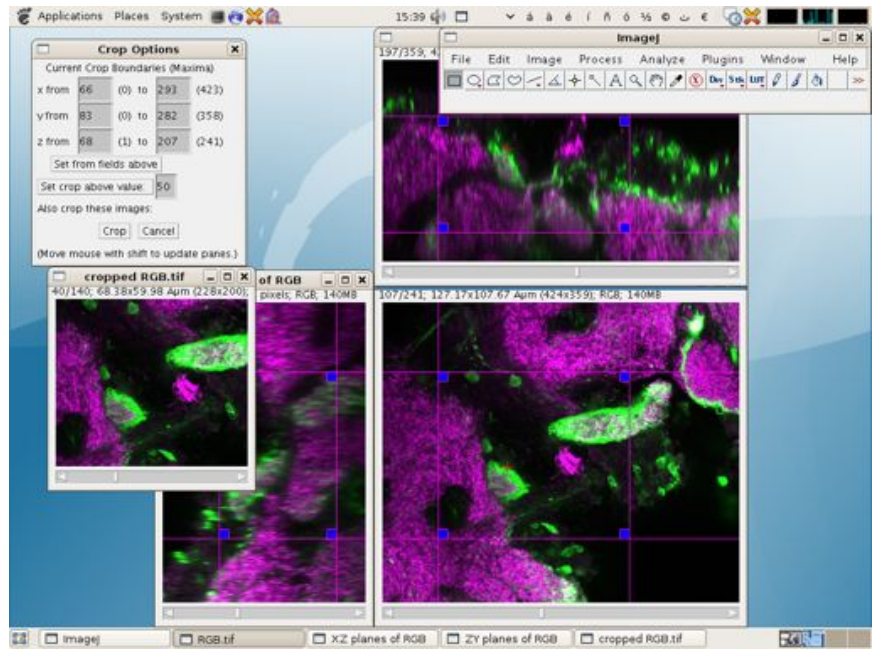

Figure 3.4:3 - ImageJ user interface[20].

## CellProfiler

CellProfiler [75] is free open-source software designed to enable biologists without training in computer vision or programming to quantitatively measure phenotypes from thousands of images automatically. It was written in Python and Matlab and is available for Windows, Mac and Linux. At the time of this thesis' writing, it supports ROI detection, ANN and SVM classification, morphological mathematic, cell counting and image pre-processing methods. A user interface is also provided.

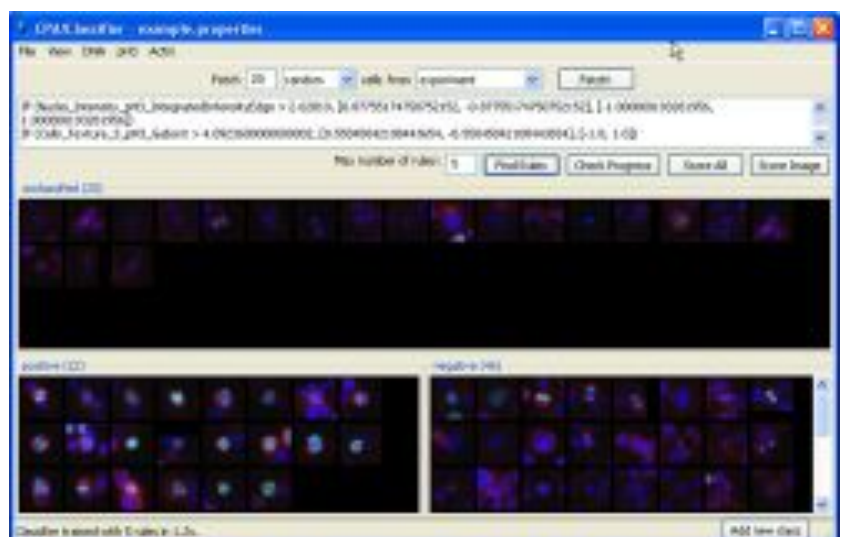

Figure 3.4:4 - CellProfiler user interface[21].

---

**CellNote**

CellNote is a software tool designed to count cells or subcellular structures in microscopic images, with the help of a computer mouse or touch input device. It allows a much more reliable and precise image annotation when compared to the classical counting performed directly under a microscope. Its ability to associate sub-objects to parent objects makes it a useful too for cellular infection level computation or polinucleated cell counting. It provides a platform for image organization, data storage and automatic spreadsheet report generation. Like the CellProfiler and ImageJ software, it also comes with a minimalistic, yet versatile, user interface.

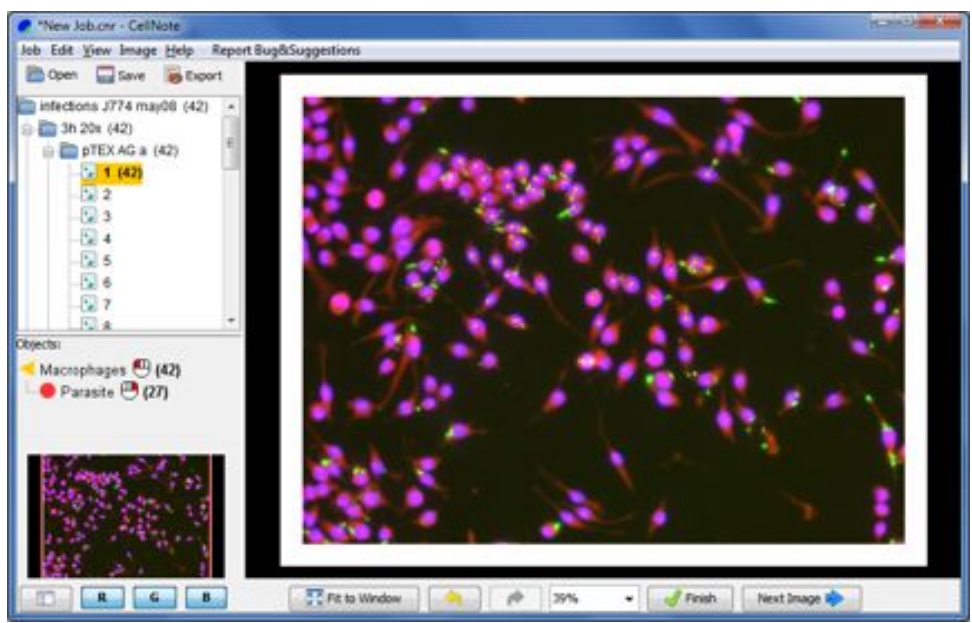

Figure 3.4:5 - CellNote user interface[22].

---

# Chapter IV

# Automatic Analysis of Microscopy Images

In chapter IV, we will describe the system architecture and its workflow. For this purpose, the chapter is divided into several sub-sections, one for each processing step. In these sub-sections, in-depth descriptions of the more complex processing steps are given, as well as critical analysis on the methods chosen for their implementation.

## 4.1 Processing Framework

This sub-section provides an overall description of the developed framework. During the development of this framework, modularity was a characteristic deemed as crucial. While the framework was meant for the implementation of a robust automatic method for the processing of fluorescence microscopy images, it should also set a standard for their analysis and thus, should provide enough adaptability for its application to other image types. The framework architecture is illustrated in Figure 4.1:1.

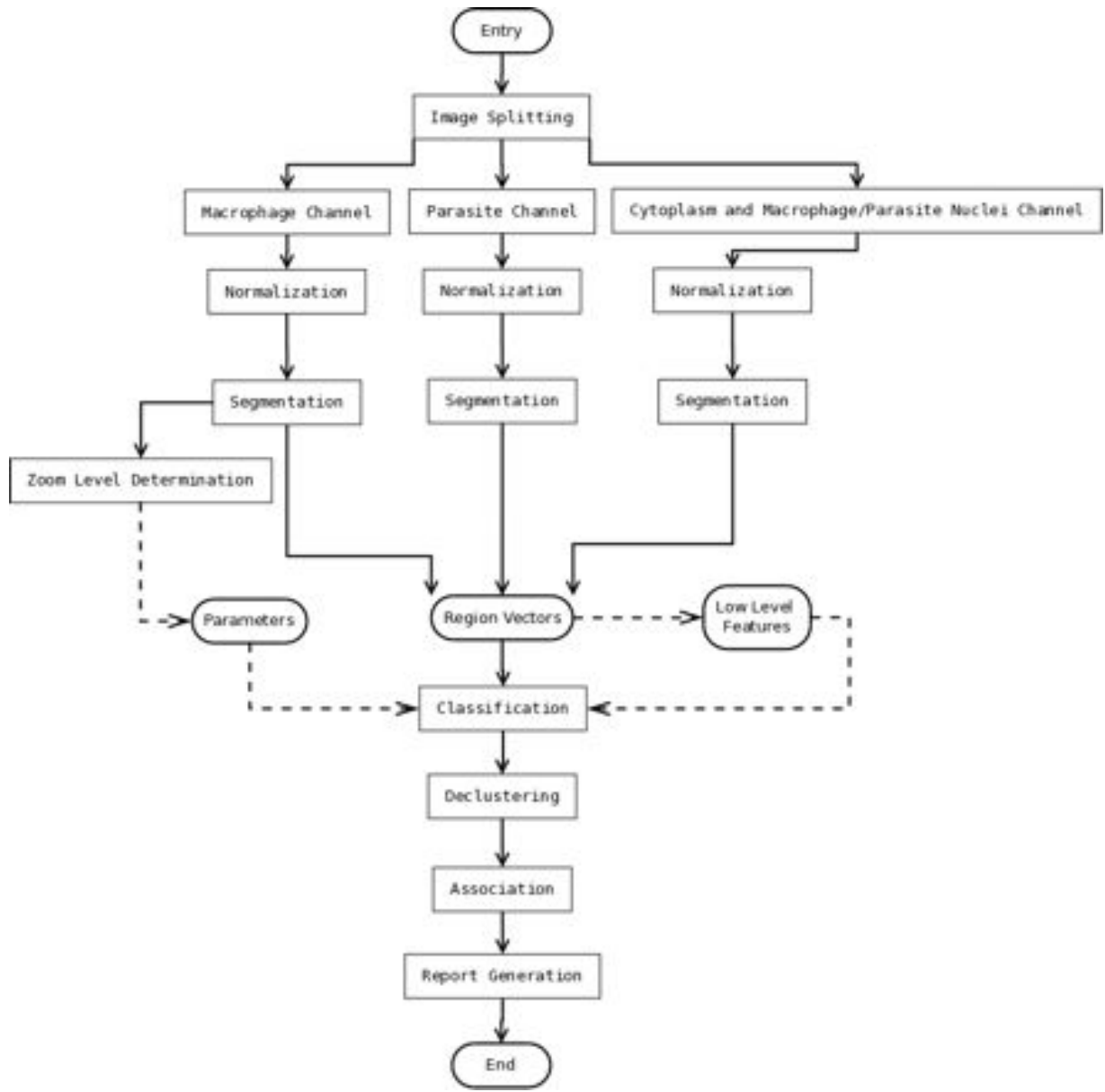

Figure 4.1:1 - Developed framework architecture.

The framework receives any type of (supported) image $f$ in the entry step. Said image is then passed on to the image splitting step, where it's processed by an image parser. This parser determines the correct image loader and then splits the loaded image into its various channels: $f_{cell}$ ; $f_p$ ; $f_{cyt}$. Each image channel then enters the pre-processing stage, described in section 4.2, and proceeds to the segmentation step. Here, the image is binarized and divided into a vector of connected regions. The regions in the cellular channel are also used to determine the zoom level the image was taken at, setting the rule set for the rule-based statistical classifier (more on this in section 4.5). Also in the segmentation step, each region vector is analysed and low level feature (LLF) vectors are computed; one per region. The next step is to classify each region; for this purpose a rule-based statistical classifier and a machine learning classifier, in conjunction with a voting system are used. Regions are then split into several sub-regions, according to the prediction made in the previous step, through a mixture model, assuming a Gaussian distribution. Finally, parasitic and cellular regions are paired in the association step and the final infection statistics and consequent report are generated in the report generation step.

In the following sections, 4.2 through 4.8, the main processing steps are discussed and their detailed architectures presented.

## 4.2 Image Splitting

Microscopy images usually come in one of two categories. Conventional image formats (such as bmp, png, gif, jpeg or tiff) or proprietary formats, such as the ZVI image format. ZVI was developed by the Carl Zeiss [76] company and is widely used in many of the company's optical and fluorescence microscopes. This format stores the image in RAW format (i.e. without any kind of compression), as well as a set of details about the image capture conditions, so that reproducing the experiment can be more easily done. These details include:

- Acquisition date;
- Microscope settings;
- Exposure values;
- Blade coordinates;
- Contrast (fluorophores) used;
- Image intervals;

The image splitting step considers both the aforementioned categories, by using a parser to chose which image loader to use. For bmp, png, gif or jpeg image formats a generic loader from Java's internal library was used. For ZVI image formats the open-source Bio-Formats library [77] was chosen. The architecture can be observed bellow (Figure 4.2:1).

It should be noted that the generic image loader always returns an image split into three channels (red, green and blue). The same does not happen with the ZVI loader library and any number of channels may be returned, one for each fluorophore used (in our case, either two, or three).

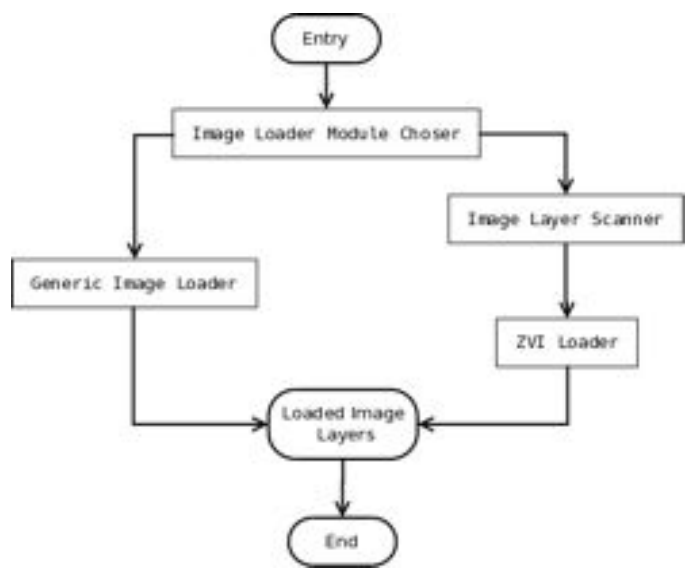

Figure 4.2:1 - Image splitting step architecture.

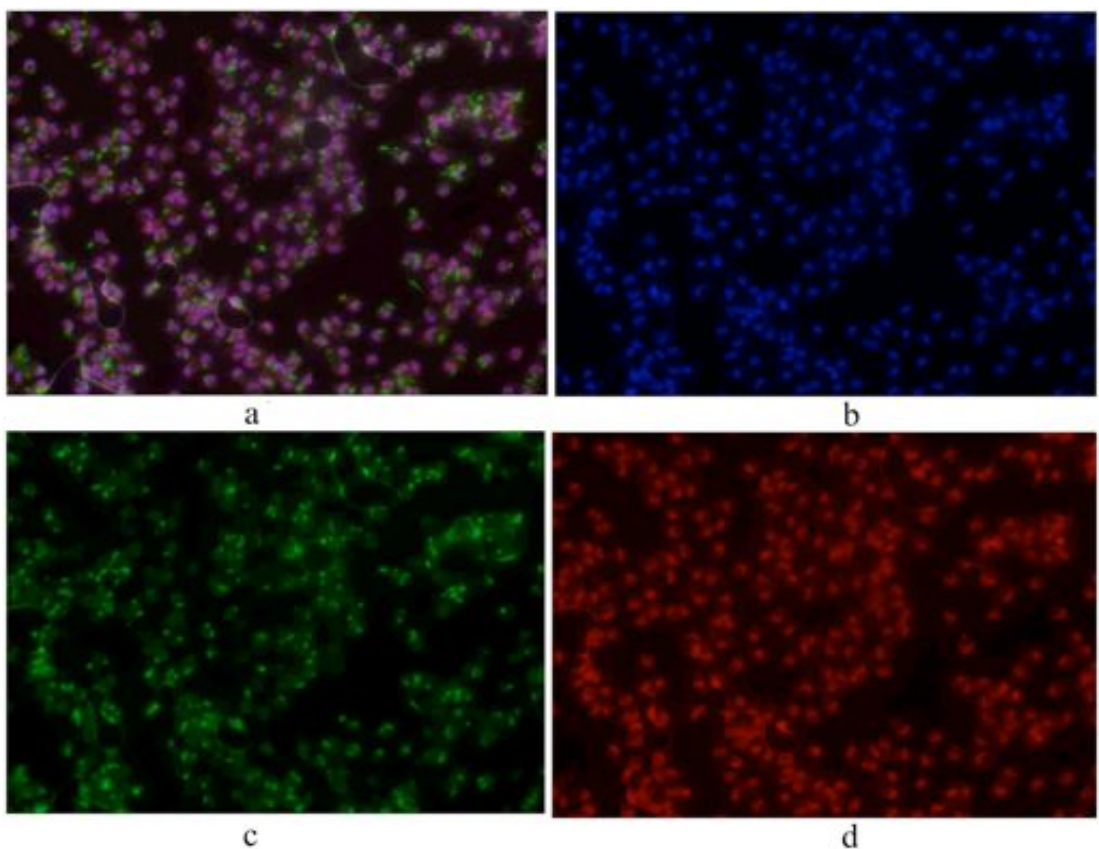

Figure 4.2:2 - a) Contrast stretched image from (same as Figure 4.3:1 - intensity differences are due to different display settings), b) Macrophage nuclei channel (blue), c) Parasitic channel (green), d) macrophage nuclei and cytoplasmic channel (red). Note that in b) and c) cytoplasmic regions are present, along with macrophage and parasite nuclei, respectively. This is due to the DNA to which each fluorophores binds itself.

## 4.3 Pre-processing

The pre-processing step is the denoted as the reunion of all the normalization operations depicted in the Figure 4.1:1. This step aims to overcome the noise produced in the image acquisition process and photobleaching effect. For this purpose, a channel independent contrast stretching operation was employed, with good results, effectively eliminating these issues, as shown by the results presented in section 5.2.

In microscopy images the spatial variations in illumination levels are normally a concern but, fortunately, this was not the case, so there was no need to apply a normalization map. Another typical issue that arises in these images is a low signal to noise ratio (SNR), which requires an additional low-pass filtering step. While this step reduces the noise ratio, it also reduces the amount of information available to make further decisions. We experienced some noise in most of the images, but the segmentation stage proved robust enough to deal with the observed amount so it was our opinion (based on an empirical observation of over 50 images) that the cost/benefit trade-off of a low pass filtering step was not good enough.

The contrast stretching method was chosen in detriment of histogram equalization because, even though the change observed is less pronounced, the applied transformation function is linear i.e. it merely enhances the image's visibility, while maintaining the original intensity distribution information. In our approach the 5th and 95th intensity percentiles were used, to avoid outlier influence.

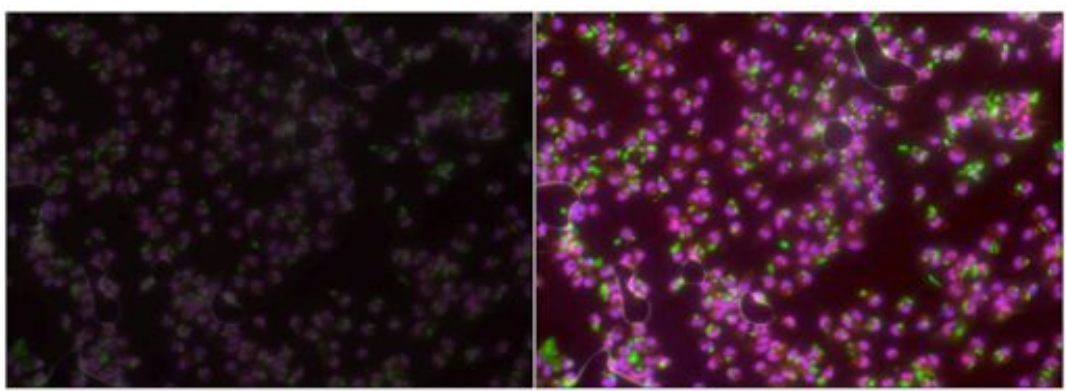

Figure 4.3:1 – Contrast stretch operation example. From left to right: original image, contrast stretched image.

## 4.4 Segmentation

In the segmentation step (as depicted in Figure 4.4:1), the image is binarized, using Otsu's Method (section 3.4.2) and then proceeds to a connected component analysis (also section 3.4.2) algorithm, where the region vector is computed.

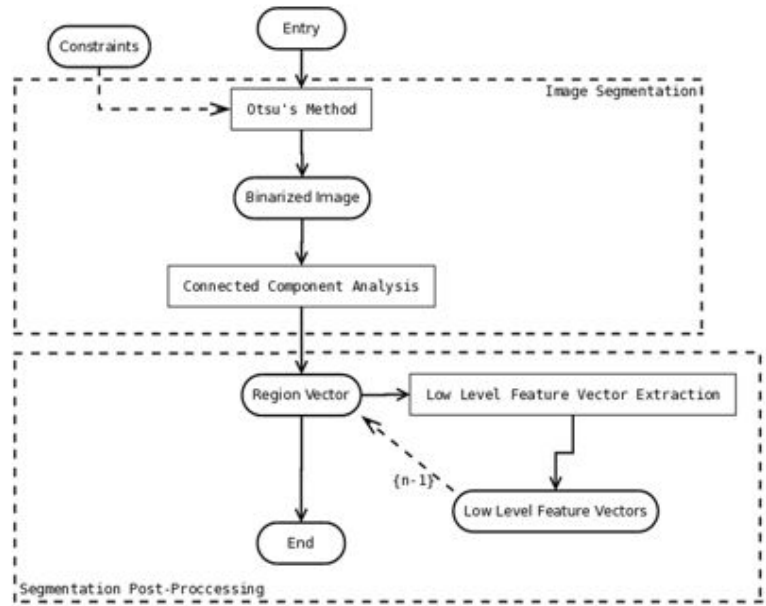

Figure 4.4:1 - Segmentation step & feature extraction architecture.

The main reason behind the choice of Otsu's Method was its applicability to this specific problem. This technique assumes that each image has a foreground and background class, which are represented by the pixel intensity values. In other words, it assumes that the image's histogram is bimodal, which is exactly what happens in our case (see Figure 4.4:2). Due to this fact, Otsu's Method performs an adequate image segmentation, without the need for any parameter fine-tuning, which is a strong pre-requisite for an automatic system. It should also be noted that other popular methods (mean shift, watersheds) were tested. These were, however, rejected due to either not being non-parametric, requiring additional methods to process their outputs (that in turn also weren't non-parametric) or simply providing better segmentation results. The method's low temporal and spatial complexity also contributed to its choice.

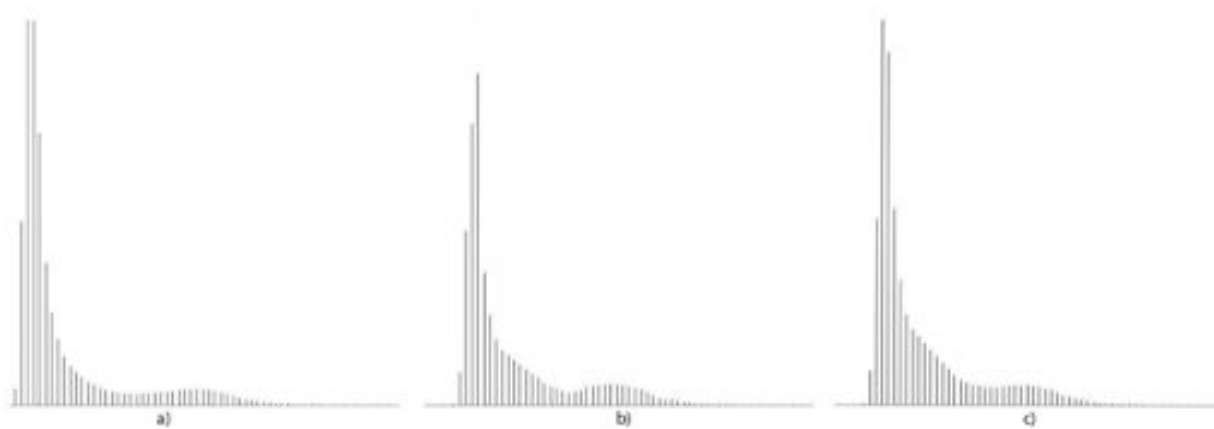

Figure 4.4:2 - Average RGB histograms from 20 fluorescence images. a) Cell nuclei channel (blue); b) Parasitic channel (green); c) Cytoplasmic channel (red). A bimodal distribution can be observed in all three histograms.

We would like to note that the segmentation results were not always ideal. This was influenced by two factors. The first, and lesser of these, were out of focus cells/parasites, that did not appear saturated enough to be considered above the segmentation threshold. This issue revealed itself not very relevant to the results because these out of focus cells and parasites were a rare occurrence, and biomedical researchers do not count them in their manual annotations anyway. The second factor was even more rare, but posed a more interesting problem. Around one for each 200 images, by our calculations, presented trimodal distributions in either (or both) the macrophage or parasite channels. This violated our previous assumption and caused a non-optimal threshold to be chosen. These were abnormal images were the cytoplasm had very low contrast and did not smoothly climb to a maximum value (the value presented by the nuclei, as this is were the DNA has higher concentration), resulting in three peaks: one for the background, another for the cell/parasite nuclei and a third one for the cytoplasmic maximum value.

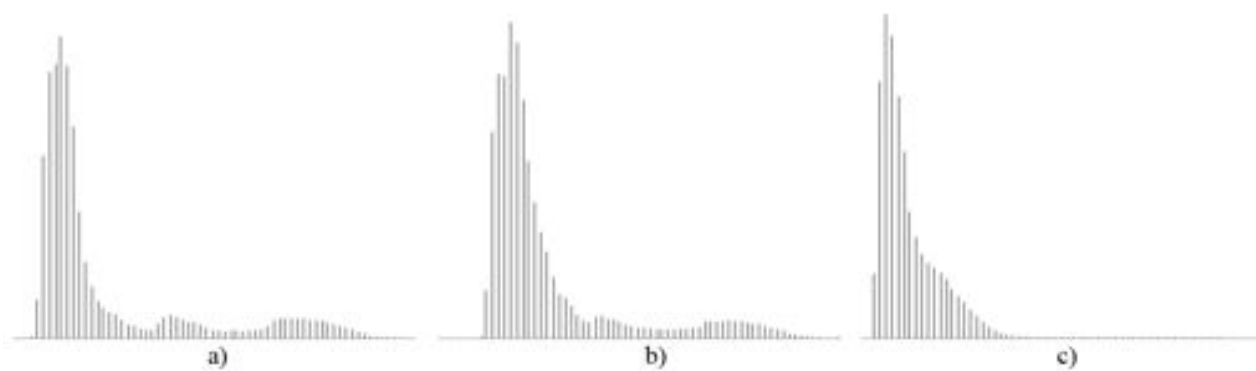

Figure 4.4:3 – RBG histograms for a fluorescence image presenting trimodal distributions in the macrophage and parasite channels. a) Macrophage nuclei channel (blue). b) Parasite nuclei channel (green). c) Macrophage/parasite nuclei and cytoplasm channel (red).

This issue was easily solved by the introduction of a histogram monotony analysis routine, which identifies the number of local maximums (peaks) present. This routine was defined over a large sliding window, so as to avoid local value fluctuations to be considered peaks. Once estimated the distribution type of each channel the segmentation could be adequately done by using a $n$-level Otsu's Method (also known as multi-Otsu), with k=$p_n$, $p_n$ being the number of peaks. The selected number of levels depends on the number of identified valleys by a monotony analysis routine. This adaptation was implemented within the method, so it is not represented in the system architecture schemes.

Since Otsu's Method has an iterative nature, additional levels represent an exponential increase in its time complexity. Following previous work, an adaptation of the proposed method in [78] was implemented. Our adaptation consisted on extending the original multi-Otsu's method with additional constraints regarding the estimated valley locations of the histogram. These can then be used to limit the search for each threshold to a specific region of the histogram. Such constraints were possible due to the images' consistent nature and resulted in a reduction from exponential to polynomial time complexity, for these particular images. Figure 4.4:4 illustrates the binarization output of our method in a highly cluttered image.

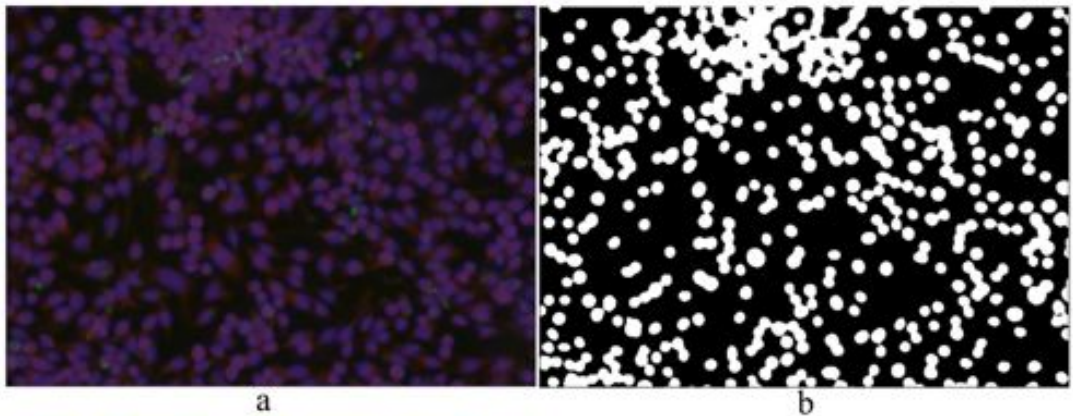

Figure 4.4:4 - Binarization output of a highly cluttered image. a) Original image; b) Binarized image, after pre-processing.

After the binarization step has been performed we proceed to a Connected Component Analysis. The traditional iterative two-step algorithm was implemented, although we provide a recursive version (which we advise against in larger images, due to obvious stack issues). A 4-point neighbourhood was used, in order to minimize the number of touching regions given the cluttered nature of the images. This procedure outputs a vector containing all of the identified regions, each with a unique id and concludes the image segmentation component. With the regions vector we now have gathered the necessary conditions to classify each macrophagic/parasitic region according to the number of nuclei contained within it, while disregarding all other irrelevant information for this task.

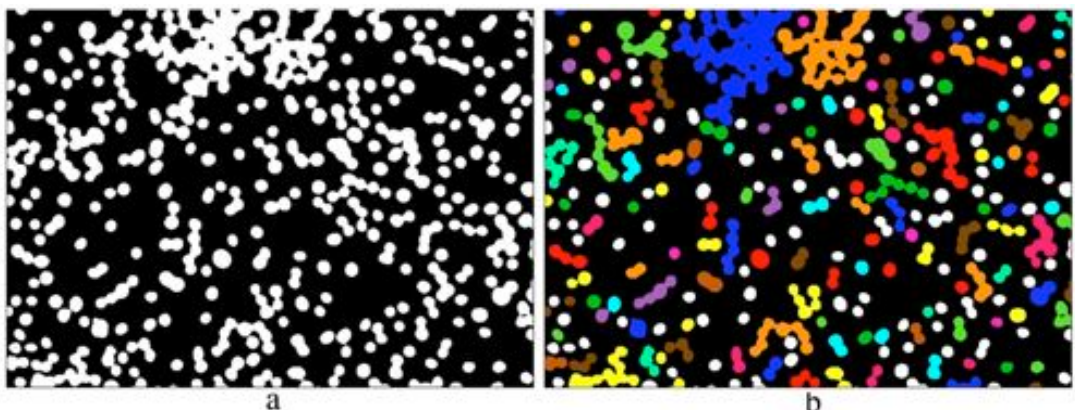

Figure 4.4:5 - Labelled image by the CCA routine. a) Binarized image from Figure 4.4:4; b) Random colour-coded visualization of the labelled image.

## 4.5 Feature Extraction

The feature extraction step (depicted as segmentation post-processing component in Figure 4.4:1) is comprised of a low-level feature extraction procedure and subsequent association of each feature vector to its respective region. These features attempt to describe characteristics that will help the classification step determine if a region has multiple nuclei (area and shape descriptor). The centre of mass, contour and bounding box are meant to aid the association step with the region's location/occupied regions of the image. Amongst the computed features are:

- Area;
- Shape descriptor;
- Centre of mass;
- Contour (unused);
- Minimum bounding box;

The feature vector can thus be defined as:

$$F_i = [a \mid cm \mid bb \mid sd] \qquad\qquad Eq.(21)$$

Where $a$ is the area, $cm$ is the centre of mass, $bb$ is the minimum bounding box and $sd$ is the roundness factor described by a 8-connected Freeman Chain Code (FCC). This shape descriptor classified each region as either circular/elliptical or other in a scalar value, by analysing the sequence pattern of codes stored in the FCC. The area and shape descriptor features were used in the classification step and the centre of mass and bounding box features were used in the association step.

Additionally, each region has several other attributes, which are helpful for future integration with a software platform such as CellNote. We highlight the following:

- Unique identifier;

- Coordinate list of each belonging pixel;
- Height;
- Width;
- Individual representation contained within a single array (for any region specific convolution or operation);
- Type flag, for classification purposes (cell/parasite nucleus, cluster or other);
- Parasite count (if cell nucleus), for association purposes;
- Random colour code, for display purposes;

## 4.6 Classification

In the classification step, the computed regions (both macrophagic and parasitic ones) are given as inputs. These regions, along with their respective feature vectors are fed to two classifiers; a rule-based statistical classifier and a machine learning classifier. The later uses a previously trained SVM model (more on this choice in the Results and Discussion chapter). The rule-based statistical classifier is parameterized with a set of parameters and analyses each region, outputting a prediction for how many nuclei (cellular or parasitic) are present. If it classifies a region as a cluster, it is also fed to the SVM classifier, who also outputs its own prediction. The two predictions are then taken as votes and a final prediction is given by a voting system (described in detail in section 4.6.3), which takes each classifier's error margin and bias into account. The final output of this step is the estimated number of nuclei present in each macrophagic/parasitic region, which will be used as the number of Gaussian mixtures to decluster said region in the next processing step.

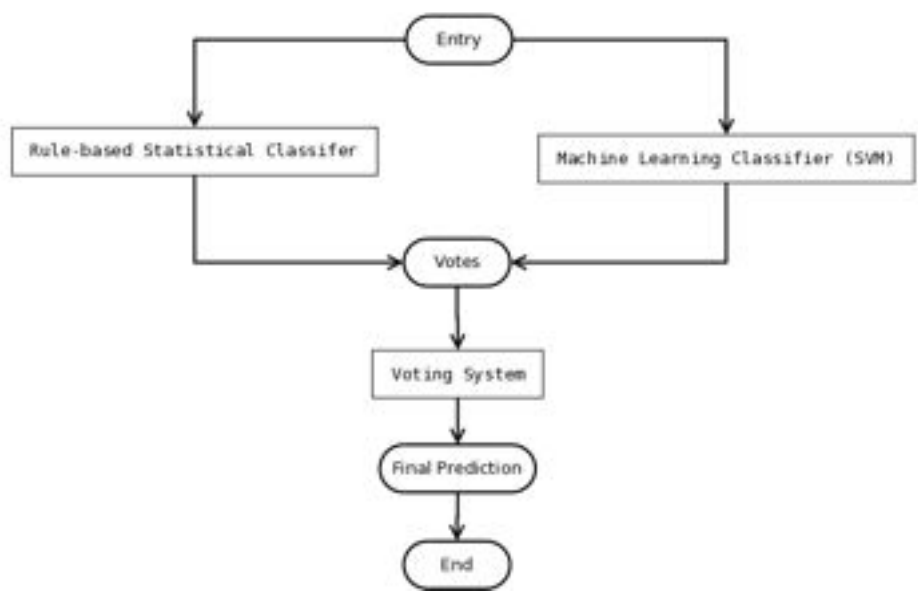

Figure 4.6:1 - Classification step architecture.

### 4.6.1 Rule-based Statistical Classifier

The principle behind this classifier is a simple one: that macrophages and parasites overlap (among them and in their individual channel) but that the overlapping amount is limited to a certain, small percentage of each ones' total area. This is due to the high depth of field observed in these images, resulting in a near perfect 2D cross-section of the 3D space within the sample.

Following this principle we hypothesised that, as cells vary in size, the area function for a region with one nucleus (from now on known as a uni-nucleic region) would present a normal distribution. In fact, this hypothesis held and uni-nucleic regions presented such a distribution with a mean value $u$ of 300 pixels (px) and a standard deviation $\sigma$ of ~48 px. Furthermore, as we assume the overlap ratio to be a low percentage of the total area, it is to be expected that multi-nucleic regions would also present a similar $\sigma$ and a mean value with $u = k * 300$, where $k$ is the number of nuclei present in the region. Since larger multi-nucleic regions are increasingly less frequent, the area values present a decreasing harmonic pattern. Although simple, we found this approach to be quite accurate (around 89% accuracy, hitting a maximum value of 98,9% when considering an error of ±2 regions as acceptable. These results will be discussed in Chapter V.

The classifier was programmed with rules reflecting this concept and was taught to ignore values outside its knowledge space (higher or lower than the maximum and lower defined bounds, respectively). The referred parameters were extracted from images with a 5 *um* zoom level, but images are typically taken at a zoom level of either 5 or 10 *um*, so an additional parameter set was extracted for 10 *u*m.

Due to the different zoom levels, for the classifier to work as intended, information about the zoom level was needed to choose the appropriate parameter set. This information is usually provided in ZVI images, but since this approach is not directed only for this image format a zoom level estimation routine was created. This routine considers the area values of each region as points in a one-dimensional feature space with two classes, which are defined by the two parameters set. The area values then vote on which class the image belongs to and a simple set of rules unties any deadlock. It is best understood through its pseudo-code:

```
routine zoomLevelDetermination(macrophagicRegions)
  zoom5Votes = 0
  zoom10Votes = 0

  for region in macrophagicRegions
    if region is not touchingImageBorders
      areaValue = region.getAreaValue()
      if areaValue < maximum5Z // u₁*2 σ₁ of the zoom level 5 parameter set
```

```
        zoom5Votes++
      else if maximum5Z < areaValue < maximum10Z //u₁*2 σ₁ of the zoom level
        zoom10Votes++                            // 10 parameter set
  if (zoom10Votes-zoom5Votes)>totalMacrophagesInImage*0.1 //if there is a
    return "zoomLevel5"                                   //large enough gap
  else if (zoom5Votes-zoom10Votes)>totalMacrophagesInImage*0.1
    return "zoomLevel5"
  else //solve tie with number of area values belonging to zoom level 10
    zoomLevel10ClusterLen = countBiggerThan(region.areas, maximum10Z)
    if zoomLevel10ClusterLen > 4
      return "zoomLevel10"
    else
      return "zoomLevel5"
```

## 4.6.2 SVM Classifier

The proposed SVM classifier uses a different approach than the aforementioned one and is aimed at estimating the number of nuclei only in multi-nucleic regions. It relies on the idea that circles or ellipses can be described as Gaussian distributions and, since macrophages and parasites have such geometry, clusters of these objects can be formulated as a GMM problem.

The classifier was built upon the following reasoning.

1) By modelling a region with $k$ mixture models we try to describe the most likely combination of parameters (of $k$ Gaussian distributions) produced by a given region;

2) Simply with this set of parameters is impossible to tell if the chosen $k$ was correct and no other method exists that correctly (enough) estimates the correct $k$;

3) We are, however, able to obtain a measure of how well this parameter combination (the built model) fits the dataset: the log-likelihood ratio;

4) Q: What happens if we start with $k$ =1 mixtures and iteratively increment this $k$ up to infinity? A: We obtain a sigmoid function, such as the one in Figure 4.6.2:1;

5) Q: Why? A: Because at first we are able to describe the dataset much better by introducing new mixtures, but as we approximate ourselves from the correct $k$ these improvements decrease. They do, however, still occur at each new introduced $k$ and only stop when each data-point is assigned to an individual mixture, at which point the log-likelihood function stalls;

6) So, if the improvement rate is described in the log-likelihood function and it is known where the correct $k$ is approximately situated, it should be possible to model a function that is able to accurately predict this $k$ for any given dataset.

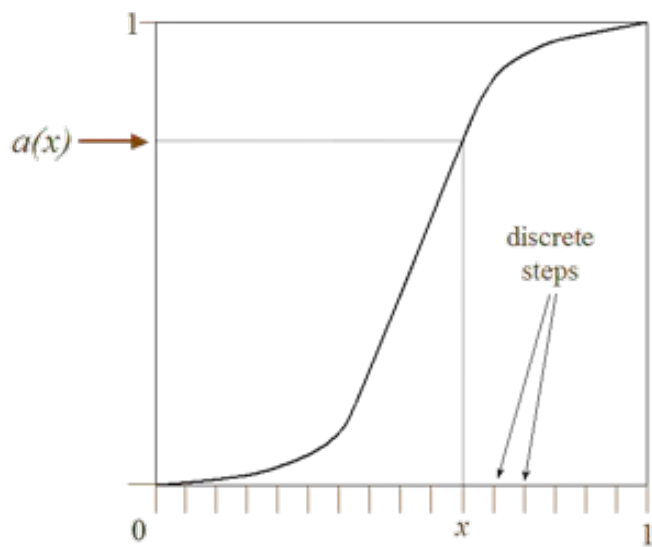

Figure 4.6.2:1 - Standard sigmoid function with discrete steps.

Based on these ideas, we extracted the log-likelihood (LL) values, as well as its first and second order derivates from approximately 150 regions, labelled them manually and used them as a training set for several ML classifiers (which will be discussed in the results chapter). Ultimately, a SVM model was chosen; as it proved itself superior to other methods, achieving an overall 85% accuracy result in a sequential split validation. We consider this accuracy percentile quite satisfactory, given the small size of the training test. The classifier's accuracy is certain to improve over time, as it is given more training data. We expect this will happen when used in a semi-automatic fashion, once integrated with our CellNote platform.

### 4.6.3 Voting System

This processing step aims at reconciling the classifiers' predictions, when these are not in accordance. This is necessary because a single classifier does not provide the most accurate prediction on its own. As previously stated in section 4.6.1, the rule-based classifier exhibits an accuracy of 89% and nearly 99%, when considered with an error margin of magnitude equal or lesser than 2. This, however, produces a reasonable amount of total error when faced with the, sometimes, thousands of nuclei present in a single image. The SVM classifier is generally much more accurate[23], but its error margins are, in counterpart, much wider (as can be observed in the confusion matrixes in Chapter V).

The voting system makes use of these characteristics to produce the more informed decision possible with the available data. It works by accepting the votes from each classifier and testing if they are in accordance. If so, the system does not intercede. If the votes differ, it compares them and chooses one based on the assumption that whatever the SVM classifier's vote, the correct decision does not deviate more than 2 from the rule-based classifier. In other words, if the SVM classifier's vote is within the rule-based classifier, it's better and should be used. If not, it is incorrect and the rule-based vote is the more accurate one.

---

*23 Note that the results presented for the statistical classifier refer to the classification of all regions (both uni-nucleic and multi-nucleic). The results for the SVM classifier only refer to multi-nucleic regions, hence to the harder parts of the region dataset to classify. For more information on each classifier's accuracy, please refer to the next chapter.*

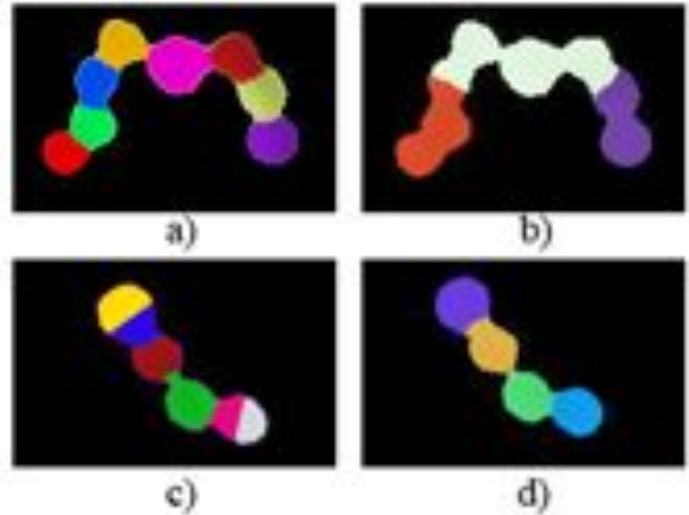

Figure 4.6.3:1 – Rule-based classifier versus SVM classifier prediction examples. a) The rule-based classifier correctly predicts the number of nuclei present in the region, while the SVM classifier provides a very wrong prediction b), more than 5 nuclei away. c) This time the rule-based classifier predicts wrongly, deviating 2 nuclei from the correct answer and the SVM classifier predicts correctly d).

## 4.7 Region Declustering

In this step, each multi-nucleic region was decomposed into the set of constituent pixels, which were provided as the available observations for the EM algorithm. The algorithm was parameterized with a minimum standard deviation of $1^{-6}$, as well as a maximum of 200 iterations. Moreover, the seeds for the centroids of each mixture were defined by the averaged centroids of a 10 fold cross-validation K-Means, to minimize runtime.

The number of mixtures used for each region was the prediction obtained from the voting system. The method performed expertly, even in the presence of large nucleic clusters made-up of a maximum of 15 nuclei. Despite this, these clusters were not taken into account, as there was not enough available data to estimate the rule-based classifier's error margins or label training data for the SVM classifier, for clusters with such magnitude.

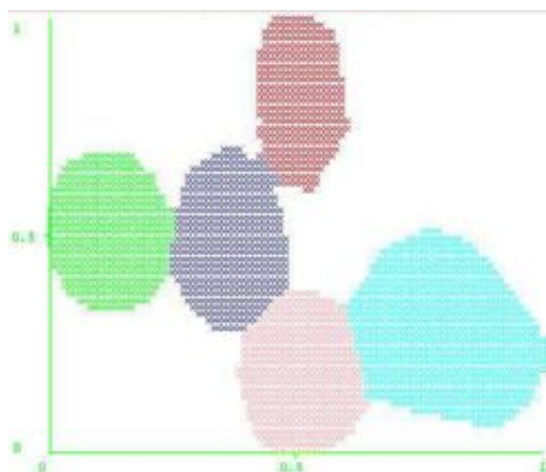

Figure 4.7:1 - Visual representation of the declustering step output. The region was processed using the aforementioned parameters and a mixture number of five.

## 4.8 Association

Two conditions are normally used to determine if a cell is infected [79]:

1.  If it cytoplasm contains one or more parasites;
2.  If a parasite is within a certain radius (~50% of the average cell radius).

Some use both conditions, while others use only condition number two. For this reason, two routines were defined in the association step; one for each of the described association methodologies. We found that, due to the cluttered nature of the images, the two association methodologies resulted in (some) changes in parasite association, but did not greatly influence the final results.

Macrophages and parasites cytoplasmic membership was trivially computed via the region vectors. The radius was taken from the zoom level parameter set chosen and calculated with the Euclidean distance from and to the centres of mass.

## 4.9 Report Generation

The final step computes the infection ratios and outputs them, along with other relevant information to a text file, located in the same directory as the target image. The report includes the following data (although a verbose mode, which is not described here, is also available):

*   Image name, with complete path;
*   Report generation date;
*   Number of detected regions (macrophagic and parasitic);
*   Number of uni-nucleic regions (macrophagic and parasitic);
*   Number of multi-nucleic regions (macrophagic and parasitic);

- Total number of counted macrophages and parasites;
- Classifier synchronization rates for macrophages and parasites (percentage of time both classifiers agreed on their vote);
- Overall infection ratio;
- Average parasites per infected macrophage;
- Average parasites per total macrophage;

An example of such a report:

```
D:\Thesis\final data\dataset 1\FS554_5uM_CS3_3_2.zvi
Report generated on: 23 - 04 - 2011 @ 23:47

Macrophagic regions: 486
Uni-nucleic macrophagic regions: 373
Multi-nucleic macrophagic regions: 29

Parasitic regions: 235
Uni-nucleic parasitic regions: 192
Multi-nucleic parasitic regions: 22

Total counted macrophages: 446
Total counted parasites: 225

Classifier synchronization rate (macrophages): 99,3318%
Classifier synchronization rate (parasites): 100%

Overall infection ratio: 0,30941704

Average parasites per infected macrophage: 1.7753624
Average parasites per total macrophages: 0.5493274
```

# Chapter V

# Results and Discussion

Chapter V presents the results obtained by the proposed method. The system evaluation provides analyses for the segmentation, rule-based statistical classifier and SVM classifier before presenting the system's final results. For this evaluation, two datasets were used; dataset A ($DS_A$) and dataset B ($DS_B$). Three biomedical researchers manually annotated both these datasets. This was used to compare our method's performance to a human.

Due to various different experimental conditions that affected the cell and parasite's sizes, the parameters sets for the rule-based classifier were slightly tuned for each one (under 10% of their total values). We consider it possible that a semi-automatic approach would be the most correct approach in such cases, where the image is too complex for a full-automatic processing. It could be adequate to implement a slider mechanism to tune each parameter set, which the user could adjust in real-time for each experiment or images considered complex enough, as he choses.

## 5.1 Materials

As previously mentioned, two separate datasets were used to validate our work. Both these datasets consisted of groups of 12 fluorescence images (totalling ~9,000 cells and ~7,000, according to the manual counting data). DA presented images with a high clustering index, putting much of the strain in the classification and declustering steps. DB presented images with a lower clustering index, but in very dim or out-of-focus conditions, naturally straining the segmentation step. Both datasets contained assorted with different zoom levels, so our zoom level estimation step could also be tested. It performed adequately, with a 100% correct hit ratio.

For the determination of infection indexes, macrophages (obtained by peritoneal lavage of mice) were seeded in 24-well plates and infected with *Leishmania* at a parasite to macrophage ratio of 10:1. One to 3 days after infection, cells were washed, fixed, permeabilized, and incubated with a specific anti-*Leishmania* antibody. The secondary antibody was the green-fluorescent Alexa Fluor 488 anti-rabbit IgG (from Molecular Probes). The preparations were additionally incubated with the blue-fluorescent DNA stain 4',6-diamidino-2-phenylindole (or DAPI) and propidium idodide (PI). The latest fluoresces in red and simultaneously detects macrophages' DNA (in the nuclei) and RNA (in the cytoplasm). Images were acquired with an Axiocam MR ver.3.0 camera coupled to an AxioImager Z1 microscope, using the Axiovision 4.7 software (all from Carl Zeiss, Germany).

For our system to be usable in a real world scenario, it should be able to perform the annotation process as well and as coherently as a biomedical researcher. To measure this, both datasets were manually annotated, in separate, by three biomedical researchers. In these annotations, each macrophage and parasite were individually identified. From these three annotations, a series of statistics were extracted (for each image):

- Total number of macrophages;
- Total number of parasites;
- Total number of infected macrophages;
- Infection ratio (infected macrophages /total macrophages);
- Average parasites per infected macrophage;
- Average parasites per total macrophages;

In order for our system's output to be considered correct it must be within a certain deviation from the manual annotations. To determine this deviation manual annotations were modelled as a normal distribution with $u = (x_1 + x_2 + x_3) / 3$ and $\sigma = \sqrt{\Sigma(x_i - u)^2 / 3}$, where $x_i$, $i=1..3$ were the manual annotations of each biomedical researcher. The algorithm's output is considered within this distribution if it does not deviated more than $\pm 2\ \sigma$ from $u$. While sub-sections 2 through 4 evaluate the targeted stress points, the final results are presented in sub-section 5 of this chapter.

## 5.2 Segmentation Results

In order to determine the segmentation step's accuracy, we measured the total number of macrophages and parasites detected versus the number present in each image. Since in our work the segmentation step is not meant to perform the whole identification process for each nucleus, multi-nucleic regions were considered well segmented. A region was considered ill segmented if: a) it was not detected or b) if it was detected but its geometry was not correctly identified (i.e. it was confused with image noise or cytoplasm).

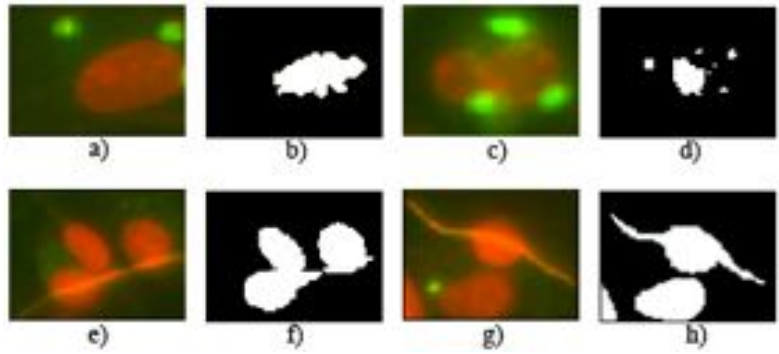

Figure 5.2:1 – Examples of ill-segmented regions (incorrect geometry).

Tables 1 and 2 present the segmentation accuracy (Eq. 22) for macrophagic and parasitic regions on both dataset A and dataset B. These results are given in percentage of the total number of macrophages and parasites present and the detection number by the method. The ground truth is given by the most complete annotation available in the dataset (i.e. the most aggressive one, among the annotations provided by the 3 biomedical researchers).

$$A(I) = \Sigma \ correct \ / \ \Sigma \ total \hspace{3cm} Eq.(22)$$

|  | Macrophages | Parasites |
|---|---|---|
| Segmentation total | 3916 | 5257 |
| Ground truth | 4025 | 5572 |
| Accuracy (%) | 97,29 | 94,35 |

Table 1.Macrophage and parasite segmentation results for dataset A.

|  | Macrophages | Parasites |
|---|---|---|
| Segmentation total | 4813 | 1832 |
| Ground truth | 5034 | 1981 |
| Accuracy (%) | 95,60 | 92,50 |

Table 2.Macrophage and parasite segmentation results for dataset B.

As can be seen in both tables, the segmentation accuracy is fairly high, ranging from a 97 to 92%. Parasitic detection is always lower than macrophage detection, by 3 to 4% in average, due to the higher contrast fluctuations found in the parasitic channel. As would be expected, the segmentation accuracy is lower in dataset B.

## 5.3 Classification Results

As dataset B possesses almost no testing data for either classifier (due to its low clustering index), both classifiers were tested with data from random regions in each of dataset A's images. Care was taken so that the data was as equally as possible divided between each class. The examples Sub-section 5.3.1 presents the results for the rule-based statistical classifier and sub-section 5.3.2 the results for the machine learning classifier.

### 5.3.1 Rule-based Statistical Classifier

This classifier was evaluated by comparing its prediction with our ground truth for each class defined in the parameter sets (detailed in Table 3). Different zoom level regions were used, but since the differences in their area values depend solely on the zoom multiplication, they were all included in the same test. This enabled us to obtain more accurate results with the same number of regions, due to the higher number of samples. No distinction was made from cellular or parasitic regions.

| Class | 0 Error Margin (correct) | ±1 Error Margin | ±2 Error Margin | ±3 Error Margin |
|-------|--------------------------|-----------------|-----------------|-----------------|
| 0 (noise) | 0.94 | 0.06 | 0 | 0 |
| 1 | 0.85 | 0.12 | 0.03 | 0 |
| 2 | 0.84 | 0.13 | 0.02 | 0.01 |
| 3 | 0.83 | 0.11 | 0.04 | 0.02 |
| 4 | 0.86 | 0.09 | 0.03 | 0.02 |
| 5 | 0.93 | 0.06 | 0.01 | 0 |
| 6 | 0.93 | 0.05 | 0.01 | 0.01 |
| 7 | 0.94 | 0.04 | 0.02 | 0 |
| 8 | 0.95 | 0.03 | 0.01 | 0.01 |
| 9 | 0.96 | 0.03 | 0.01 | 0 |
| 10+ | N/A | N/A | N/A | N/A |

Table 3.Rule-based statistical classifier's sensibility and specificity results. Note that the class 10+ is not included as there was not enough data to obtain a statistically meaningful result. These results were obtained from approximately 800 regions. Also, an error margin higher than ±3 was never observed.

It is clear that while the classifier is able to correctly identify regions of one or two or three nuclei, but as the number of nuclei present increases past this number its error margins quickly decrease. This is because, as the number of nuclei present in a region increases, the more its area value approximates the mean value (times the number of present nuclei) because more samples of the same distribution are being taken into account. However, we chose not to classify regions with over 9 estimated nuclei, due to two reasons.

1) We did not have enough training data to appropriately train the SVM classifier or test how the declustering step would perform in these situations;

2) Typically, biomedical researchers make a good practice in not counting regions that are very infected.

Since the number of regions with low nucleic numbers is predominant in most images, the classifier's accuracy is the weighted sum of the hit rate (0 error margin). In the used testing set we obtained the following overall results:

- Total number of regions: 793;
- Correctly classified regions: 706 (~89%);
- ±1 error margin: 44 (~5,5%);
- ±2 error margin: 11 (~1,4%);
- ±3 error margin: 4 (~0,5%);

## 5.3.2 SVM Classifier

To train the SVM classifier, a feature extraction routine was developed. This routine was used to export the data from 150 random multi-nucleic regions to an external training dataset file, which was then read, pre-processed and used to build several classifiers using Weka. The feature vectors were defined as:

$$F = [LL, FOD, SOD, A] \qquad (23)$$

With the following description:

- Log-likelihood (*LL*): values obtained by performing an EM algorithm on the region data with up to ten mixture numbers - ten coefficients;
- Discrete first order derivate (*FOD*) of the LL values - nine coefficients;
- Discrete second order derivate (*SOD*) of the LL values - eight coefficients;
- Region area (*A*) - one coefficient;

Among the tested methods, we chose to highlight only the ones that exhibited the higher accuracy percentiles. These where:

- Best First Search Tree – a tree search algorithm that explores a graph by expanding the most promising node chosen according to a specific rule [80];

- C4.5 Search Tree – an updated version of the ID3 classifier, which uses the concept of information entropy to create a decision tree [81];
- Multiplayer Perceptron Neural Network – a feed-forward artificial neural network that maps sets of input data onto a set of appropriate output [82];
- Support Vector Machine – a support vector machine with a sequential minimal optimization training algorithm [65];

The classifiers were evaluated using three solution analysis techniques: sequential splitting [82] and cross-validation [83]. The sequential splitting was performed, with splits at 66 and 90% and the cross-validation with two and ten folds. The overall results for all 4 algorithms can be observed in Tables 4 through 7. In these tables, d' and d'' denote the first and second derivates of the LL values. ArT and ArZ denote the area values where no distinction was made for the zoom levels and area values where such distinction was made (in these cases the results provided are the average of the results for both zoom levels).

| Best First Search Tree | | | |
|---|---|---|---|
| | 2 fold CV | 10 fold CV | Seq. Split 66% |
| LL-d'-d'' | 79.2 | 73.3 | 82.4 |
| LL-d'-d''-ArT | 77.2 | 77.2 | 67.6 |
| LL-d'-d''-ArZ | 72.9730 | 70.3 | 69.2 |

Table 4.Best first search tree accuracy results.

| C4.5 | | | |
|---|---|---|---|
| | 2 fold CV | 10 fold CV | Seq. Split 66% |
| LL-d'-d'' | 77.2 | 76.2 | 79.4 |
| LL-d'-d''-ArT | 75.2 | 75.2 | 79.4 |
| LL-d'-d''-ArZ | 67.6 | 56.8 | 69.2 |

Table 5.C4.5 accuracy results.

| FFNN | | | |
|---|---|---|---|
| | 2 fold CV | 10 fold CV | Seq. Split 66% |
| LL-d'-d'' | 77.2 | 73.3 | 82.4 |
| LL-d'-d''-ArT | 70.3 | 68.3 | 70.6 |
| LL-d'-d''-ArZ | 67.6 | 73.0 | 61.5 |

Table 6.Feed-forward neural network accuracy results.

| SVM | | | |
|---|---|---|---|
| | 2 fold CV | 10 fold CV | Seq. Split 66% |
| LL-d'-d'' | 78.2 | 75.2 | 85.3 |
| LL-d'-d''-ArT | 74.3 | 73.3 | 76.5 |
| LL-d'-d''-ArZ | 78.4 | 73.0 | 69.2 |

Table 7.Support vector machine accuracy results.

Upon analysing the results provided by several solution analysis techniques we made 2 conclusions regarding them:

1) The re-substitution technique provided an accurate description of how well the built model represented the training data. However, it did not represent a good indicator of how well this model would fit new, unlabelled data. Methods created resorting to this procedure tended to exhibit over-fitting, so it was not factored into the final decision;

2) Due to the low number of regions in the training data, the sequential split at 90% provided erratic results. These were also not a good indicator of the model's accuracy, since a testing set of 15 regions is not large enough to be taken as statistically relevant;

For these reasons, only the sequential split at 66% of the training set and the 2 and ten fold cross validation techniques were factored in the final decision. Out of the used training sets, the first one produced, in average, the best results for every algorithm. Out of all the algorithms, the SVM presented higher scores with this training set, produced competitive running times, revealed itself less prone to over-fitting or affected by the data's high dimensionality, so our final choice fell on this method.

All four algorithms presented similar confusion matrixes, so they are not discussed in detail (although they are available in the appendix section). The SVM did, however, show a higher tendency to greater classification errors. In any other case this would be considered as a con but, in our case, it can be used to our advantage by using these errors to rule-out the SVM's prediction and using the rule-based classifier's instead.

## 5.4 Final Results

For the final results all images from both datasets were processed through our algorithm. The data obtained from the generated report files was then compared with each of the three manual annotation available and evaluated in a similar fashion as in the segmentation results. In other words, annotation values were modelled as a normal distribution and the method considered accurate if its output did not deviate more than 2 standard deviations (Eq. 25) from the computed mean value (Eq. 24).

$$\mu = 1/N \sum_{i=1}^{N} x_i \tag{24}$$

$$\sigma = \sqrt{1/N \sum_{i=1}^{N} (x_i - \mu)^2} \tag{25}$$

$$e = \left\| 2\sigma \right\| \tag{26}$$

We would like to note that the two most common parameters extracted from *Leishmania* images are the percentage of infected macrophages and the average of parasites per infected macrophage. Our decision to not evaluate our system's performance based on these ratios was motivated by the notion that they are calculated through the total number of cells, parasites and infected cells. Evaluating the system's performance separately on these values instead provides a better understanding of its accuracy and final results. While we did not base our performance analysis on these ratios, we did evaluate them, with the algorithm always obtaining better results than the ones presented by the total number of cells, parasites and infected cells (due to error cancelation). However, these ratios are needed to judge if it our algorithm is able to substitute a human in an annotation task, so they are present in the appendix section. Tables 8 and 9 present the final results for dataset A. Tables 10 and 11 present the final results for dataset B.

|        | Macrophages (total) | Parasites (total) | Infected Macrophages |
|--------|---------------------|-------------------|----------------------|
| BR-A   | 4025                | 5572              | 2595                 |
| BR-B   | 3164                | 3556              | 1661                 |
| BR-C   | 1872                | 2983              | 1364                 |
| PA     | 4373                | 5611             | 2592                 |

Table 8.Final results for Dataset A. Legend: BR-A: Biomedical Researcher A; BR-B: Biomedical Researcher B; BR-C: Biomedical Researcher C; PA: Proposed Algorithm. [24]

---

24 BR-A/B/C: Manual annotation values.

|  | Macrophages (total) | Parasites (total) | Infected Macrophages |
|---|---|---|---|
| Mean (μ) | 3020 | 4037 | 1873 |
| Standard Deviation (σ) | 885 | 1110 | 546 |
| μ +2σ | 4790 | 6258 | 2922 |
| μ -2σ | 1250 | 1816 | 824 |
| Algorithm Error (26) | 1353 | 1574 | 719 |

Table 9.Acceptable error margins for Dataset A.

In this dataset we can clearly see a large discrepancy in the three manual annotations. BR-A was clearly the most daring researcher, counting around twice the regions BR-C counted. In comparison, BR-C was the one most aggressive when associating cells and parasites (given the amount of detected cells). These (and other) annotation discrepancies are the main contributors to the enormous standard deviation observed; nearly 30% for cells, 25% for parasites and, again, nearly 30% for infected cells. Due to these large discrepancies, our method easily fits within the defined boundaries, actually being closer to one standard deviation in total parasites and infected cells. We did not expect the images to be so difficult to annotate that the process would generate such disagreement between the biomedical researchers. In the future we plan to assemble a group of researchers with whom we will discuss their annotations, thus striving to obtain a consensus and more homogeneous annotations (and with this knowledge, better tune the algorithm).

|  | Macrophages (total) | Parasites (total) | Infected Macrophages |
|---|---|---|---|
| BR-A | 5034 | 1981 | 1015 |
| BR-B | 5446 | 2001 | 1075 |
| BR-C | 4728 | 1920 | 984 |
| PA | 5292 | 1834 | 1052 |

Table 10.Final results for Dataset B. Legend: BR-A: Biomedical Researcher A; BR-B: Biomedical Researcher B; BR-C: Biomedical Researcher C; PA: Proposed Algorithm.[25]

---

25 BR-A/B/C: Manual annotation values.

|  | Macrophages (total) | Parasites (total) | Infected Macrophages |
|---|---|---|---|
| Mean (μ) | 5069 | 1967 | 1024 |
| Standard Deviation (σ) | 294 | 34 | 38 |
| μ +2σ | 5658 | 2036 | 1100 |
| μ -2σ | 4481 | 1898 | 949 |
| Algorithm Error (26) | 223 | 133 | 28 |

Table 11.Acceptable error margins for Dataset B. Since DS$_B$ was meant as a segmentation stress test under poor illumination and exposure settings, it was expected that the number of detected regions would be lower. While the number of detected cells is within one standard deviation of the closest annotation, the number of detected parasites exceeds 3σ.

In sub-section 5.3.1 results showed that the segmentation step performed its worst for parasite detection in dataset B. Since little to no multi-nucleic regions appeared in this dataset, no fault could be attributed to the classification step and it would seem like the low parasitic detection (and subsequent error) falls upon the segmentation step. Clearly, the discrepancies seen in dataset A are no longer present in dataset B, which contributed to the low standard deviation verified, which further increased the error seen in the segmentation results.

Before presenting our final remarks on these results, we would like to point out reasons we discovered for much of the documented deviations:

- Some researchers count cells on the borders of the images. As a convention we did not do this, to minimize possible errors. In regions with low cell or parasite numbers this can affect the final annotation results (Figure 5.4:1);
- The number of counted cells (and parasites) depends on how 'daring' the researcher is. This varies greatly from person to person and from annotations performed in different dates by the same person (Figure 5.4:2);
- The association radius, while somewhat defined, still varies greatly from person to person, in time and inside each image (Figure5.4:3);
- We would also wager that researchers from different labs tend to have even different annotation habits, which would further increase the observed deviation indexes;

Pertaining dataset A, three images exceeding the two deviations in one of the counts were recorded. These images were highly out of focus or had severe under-exposure issues. Only in two of these did these deviations significantly affect the infection and average parasite per infected region significantly (i.e. causing the image to be considered badly annotated). The low parasite count is mainly due to three factors:

1) The low exposure conditions observed in this particular dataset;

2) Blurred parasites, which led the algorithm to detect fewer of these and the researchers to increase their aggressiveness, further widening their gap with the program.

3) Since the issue with identifying the parasites is detecting them (and humans can do this extremely well), it led to a small deviation between their annotations. So, even though the algorithm's deviation is not great in absolute numbers, the low relative differences between annotations still implies a large enough error;

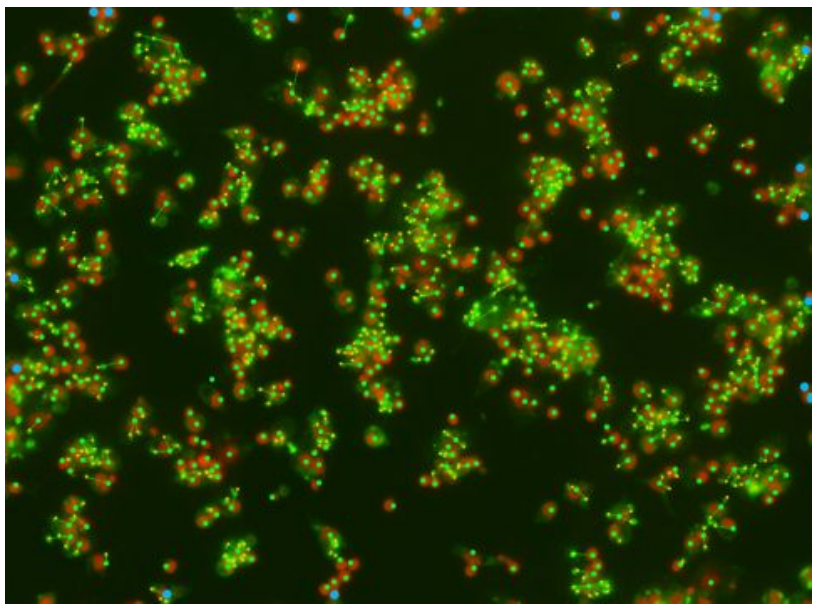

Figure 5.4:1 – Image with macrophages and parasites located at the image's border annotated (blue dots).

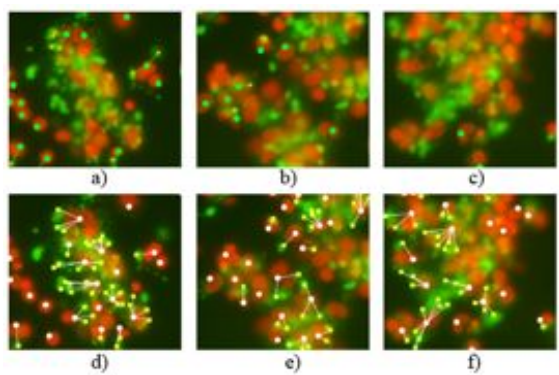

Figure 5.4:2 – Comparison between the aggressiveness of two biomedical researchers. a), b) and c) Details of a sample image annotated by researcher I. d) e) and f) Same details, annotated by researcher II. A clear difference in aggressiveness can be observed between both researchers.

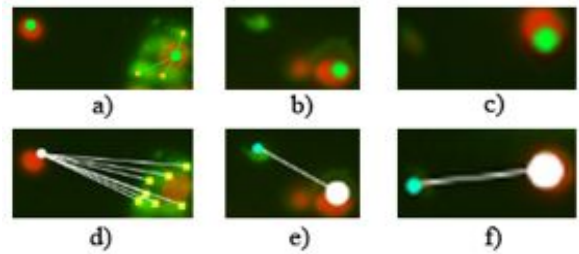

Figure 5.4:3 – Comparison between the association radius used by two biomedical researchers. a), b) and c) Details of a sample image annotated by researcher I. d), e) and f) Same details, annotated by researcher II.

Some measures to circumvent these issues are described in some detail in focuses one to three of the future work sub-section of the next chapter.

Regarding dataset B, the major discrepancies were due to cell counting errors. Some lesser errors were registered in parasite counts. Sometimes, two opposing errors can lead to correct final results. Since each count (cells, parasites and infected cells) was within the designated error margins seemed to not be the case. This was, however, the case in four occasions between the total results of individual images. In our opinion, these deviations were not due to the method's ill performance, but rather mostly due to variations in personal annotations (as they were taken across nearly a two month period).

In sum, given the adverse testing conditions and the problem's complexity, we consider the final results both acceptable and encouraging. Future work in defining semi-automatic processing for complex images and further training the SVM classifier will, undoubtedly, improve these results. We would like to conclude with some interesting details from several images' final results, depicted in Figure 5.4:4.

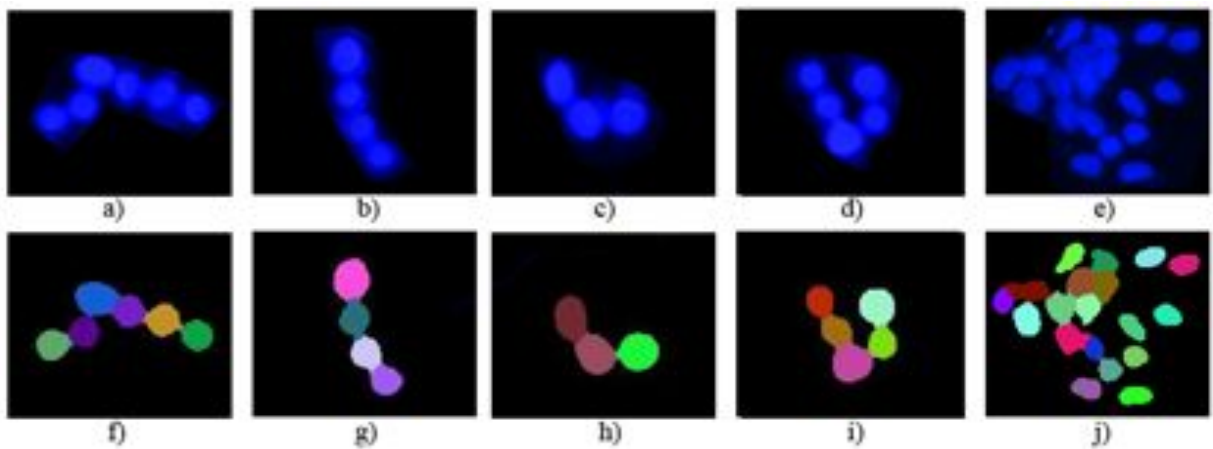

Figure 5.4:4 – Several declustering examples, chosen from $DS_A$ and $DB_B$. a) - e) Original region patches. f) - j) Respective declustered patches.

# Chapter VI

# Conclusion

In this chapter we present our conclusions on the developed work. Section 6.1 confronts the obtained results with the intended contributions. Section 6.2 describes future work that will act as a direct extension and improvement of this research.

## 6.1 Goals

The main contribution intended by this work was to develop a robust automatic computer-vision based methodology for macrophage and parasite detection and infection level determination in fluorescence microscopy imaging. Our method is robust to poor lighting conditions and high nucleic-cluttering indexes and results show that even in difficult situations, it is well within the error margins observed by biomedical researchers. Furthermore, with its future integration with our CellNote platform, the classifier will have a change to build a much larger training set and provide even more accurate results.

The second contribution we aimed at was the creation of a dataset for the testing and development of solutions for this problem. For this study a dataset of nearly 1,267 fluorescence and 623 Giemsa (which were not used) images was collected and manually classified into three difficulty settings, as well as two other categories (high cellular and high parasitic cluttering). We, therefore consider this contribution successful.

The third contribution was the creation of a standard pipeline for cellular image processing. Through chapter IV we introduced the various components of our developed framework, detailing each step and it's workflow. Not only is our proposed framework modular, as it is also and extensible and can be applied to other image types by simply changing the image loader used and editing or adding steps to the pre-processing stage.

As a consequence of our result evaluation a final, fourth contribution was made regarding the implications of different association methodologies, as well as the intra and inter-person variances in the manual annotation process of microscopy images.

In sum, we consider this work to have been a success and to have exceeded our initial expectations, providing biomedical researchers with a useful tool for the acceleration of *Leishmania* and, in the future, more general microscopy imaging-based research.

## 6.2 Future Work

Future work should have four distinct focuses. Firstly, to enhance the accuracy of the segmentation process, while maintaining its robustness. One possible approach is to adapt the existing algorithm to a sliding window mechanism, creating a local segmentation. Although Otsu's method seems adequate for this task it would be interesting to also explore other segmentation algorithms, such as mean shift, level sets or normalized cuts, especially when applying our algorithm to other microscopy image types (e.g. Giemsa images).

The second focus should be to increase the rule-based classifier's accuracy. Further studying the decreasing harmonic function, given by the area values and defining improved parameter sets is one approach. To create larger training sets, with more reliable ground-truth information and employing ML methods to model the distributions of the data values is another. Regarding the SVM classifier, the accuracy will improve over time so continued training should suffice.

A third focus is to create semi-automatic processing schemes for individual experiments or images too complex for automatic processing. This would imply the creation of parameter adapting interfaces for the segmentation threshold constraints and rule-based classifier parameter sets, as well as the association radius. The development of an algorithm to detect the complexity level of each image would also be necessary, so as to decide when to quit automatic processing and ask for user guidance (from which it could learn).

The fourth and final focus would be to apply the developed framework to the processing of other microscopy image types, such as Giemsa [85] images. This is left as an open problem to the scientific community and biomedical laboratories interested in such systems.

We expect the first two of these to be developed shortly within the scope of the CellNote and CellNote Touch [86] projects. The third is an open possibility, which could be explored in subsequent research projects. Unfortunately, time did not allow for the study of these suggestions, that while simple consume a considerable amount of time, so we hope to see them fulfilled in the future.